\documentclass{article}

\usepackage[nonatbib, final]{neurips_2019}

\usepackage[utf8]{inputenc} 
\usepackage[T1]{fontenc}    
\usepackage{hyperref}       
\usepackage{url}            
\usepackage{booktabs}       
\usepackage{amsfonts}       
\usepackage{nicefrac}       
\usepackage{microtype}      

\usepackage{bm}             
\usepackage{graphicx}       
\usepackage{algorithm}      
\usepackage[noend]{algpseudocode} 
\usepackage{caption}        
\usepackage{array}          
\usepackage{booktabs}       
\usepackage{pbox}           
\usepackage{subcaption}     
\usepackage{cite}

\title{Stochastic Adversarial Koopman Model for Dynamical Systems}

\author{%
  Kaushik Balakrishnan \\
  Ford Greenfield Labs, Palo Alto, CA \\
  \texttt{kbalak18@ford.com} \\
  \And
  Devesh Upadhyay \\
  Ford Research, Dearborn, MI \\
  \texttt{dupadhya@ford.com} 
}

\begin{document}

\maketitle

\begin{abstract}

Dynamical systems are ubiquitous and are often modeled using a non-linear system of governing equations. 
Numerical solution procedures for many dynamical systems
have existed for several decades, but can be slow due to high-dimensional 
state space of the dynamical system. Thus, deep learning-based reduced order models (ROMs) are
of interest and one such family of algorithms along these lines are based on the Koopman theory.
This paper extends a recently developed 
adversarial Koopman model (Balakrishnan \& Upadhyay, arXiv:2006.05547) to stochastic
space, where the Koopman operator applies on the probability distribution of the latent 
encoding of an encoder. Specifically, the latent encoding of the system is modeled as a 
Gaussian, and is advanced in time by using an auxiliary neural network that outputs
two Koopman matrices $K_{\mu}$ and $K_{\sigma}$. 
Adversarial and gradient losses are used and this is found to lower the prediction errors.  
A reduced Koopman formulation is also undertaken where the Koopman matrices are
assumed to have a tridiagonal structure, and this yields predictions comparable to the
baseline model with full Koopman matrices. The efficacy of the stochastic Koopman model is demonstrated on different 
test problems in chaos, fluid dynamics, combustion, and reaction-diffusion models.  
The proposed model is also applied in a setting where the Koopman matrices 
are conditioned on other input parameters for generalization and this is 
applied to simulate the state of a Lithium-ion battery in time. 
The Koopman models discussed in this study are very promising for the wide
range of problems considered.

\end{abstract}


\section{Introduction}

Many engineering and scientific problems are modeled as dynamical systems and are very
often represented as a system of partial differential equations (PDEs), which are solved using robust numerical methods such as finite difference or
finite volume methods. Such methods are, however, time consuming to be solved numerically, particularly if the state space of the system is high-dimensional. 
Thus, recasting the problem in a lower-dimensional state space and advancing the solution in time in this reduced dimensional space is
preferred as it saves computational time. Many reduced order models (ROMs) based on Proper Orthogonal Decomposition (POD)
have been developed and are widely used by the research community \cite{POD1, POD2, DEIM}.
Also of interest are ROMs based on Koopman models \cite{Koopman1931, Koopman1932} where a non-linear model is assumed to have linear
dynamics in a Koopman-invariant subspace. Here, numerical methods such as 
dynamic mode decomposition (DMD) \cite{Schmid2010, Kutz2016} are popular. 
In the Koopman \cite{Koopman1931, Koopman1932} family of algorithms for modeling non-linear dynamical systems, 
the system dynamics is assumed to be linear in a sub-space and the solution is evolved in time in this sub-space, after
which it is mapped back to the higher dimensional physical space.
Several research works have renewed interest in Koopman models in recent years,
as evidenced by \cite{Mezic2004, Mezic2013, Arbabi17, Lusch2018, Salova19, Balakrishnan20}.

Function approximation of data is critical to such models and to this end
Deep Learning is a natural choice since neural networks are not only universal function approximators \cite{Hornik1989, Hornik1991}, 
but are also robust to represent high-dimensional data.
For these reasons, Deep Learning is widely applied in many areas, inter alia, computer vision, robotics, audio synthesis, natural language processing, etc. 
For example, neural networks-based autoencoders have been developed for dimensionality reduction \cite{AE} and are 
solved using the Backpropagation algorithm \cite{Backprop}. Generative models such as Generative Adversarial Networks (GANs) \cite{GAN} have  
demonstrated excellent performance in computer vision, speech synthesis, and natural language processing.  
GANs are the new gold standard in deep generative models and many flavors of GANs have been developed in these fields with customized tasks.

Many engineering problems 
can be tackled with the blending of deep learning and classical Koopman theory-based approaches, as
evidenced by \cite{Takeishi2017, Yeung2017, Lusch2018, Morton2018, Morton2019, Balakrishnan20}.
In \cite{Lusch2018, Balakrishnan20}, an auxiliary neural network was trained to obtain the Koopman eigenfunctions, whereas in \cite{Takeishi2017, Morton2018}
a least squares approach was used. In \cite{Yeung2017}, a global Koopman operator was learned as part of the neural 
network optimization which, however, may not be possible in other engineering problems.
 
Following the developments in \cite{Balakrishnan20}, we extend the solution procedure for dynamical systems using a stochastic 
latent embedding space, and the Koopman operator applies in this stochastic space. Specifically, the Koopman operator
applies on the probability distribution of data and advances linearly in time, thereby learning a distribution of possible outcomes. 
In other words, if $z$ is the latent embedding and 
$p(z)$ is its probability distribution, past studies \cite{Takeishi2017, Lusch2018, Morton2018, Balakrishnan20}
apply the Koopman operator directly on $z$; in this study, we apply the Koopman operator on $p(z)$ instead. 
A similar idea was used recently in \cite{Morton2019}, but the authors took a variational approach, whereas we 
assume data in the latent space is Gaussian and apply the Koopman operator directly on the inferred
mean and standard deviation of the data. Specifically, we consider two Koopman matrices, $K_{\mu}$ and $K_{\sigma}$, which 
advance the mean and standard deviation of the Gaussian distribution $p(z)$.
Real-world data can have uncertainties, noise, chaos, etc. \cite{Mandic08, Strogatz00, wiggins03} and so using a stochastic model to capture the data 
distribution can help in certain settings, and this
forms the primary motivation for us to investigate the use of the Koopman operator on data distribution and not on the data sample per se. 
For instance, in chaos and bifurcation theory (e.g., pitchfork bifurcation), the solution of a dynamical system can change drastically when a minor perturbation is added
to the system, and so being able to predict a distribution for the next states of the dynamical system can be richer than treating the problem deterministically.
In Reinforcement Learning literature, others have also demonstrated that the use of stochastic representation learning 
augments the overall predictability \cite{SLAC, Ha2018, Nachum2019}, as the uncertainty associated with data is also learned, thereby one can overcome a
``representation learning bottleneck'' \cite{SLAC}. 
  
Some of the machinery used in this study extends ideas from \cite{Balakrishnan20}.
An autoencoder is used whose encoder maps data to the latent space where it is represented as a Gaussian distribution. 
An auxiliary network is also used for generating two Koopman matrices, $K_{\mu}$ and $K_{\sigma}$, which operate on the mean and standard deviation of
the data distribution in the latent space. The decoder subsequently maps the data back from the latent space
to the physical space. A GAN discriminator is also used and this is found to robustify the predictions, as we will demonstrate.  
A series of test problems are considered for the analysis ranging from chaos, fluid dynamics, reaction-diffusion problems, and Lithium-ion battery modeling.


\section{The Koopman operator for dynamical systems}

We consider discrete in time ($t$) dynamical systems of the form: 
\begin{equation}
x_{t+1} = F\left(x_t \right ),
\end{equation}
where $x_t$ is the state of the system at time $t$, which is high dimensional typically, $x_t \in \mathcal{R}^N$ and $F$() is a
non-linear function that describes the dynamics of the system (e.g., the Navier-Stokes equations of fluid dynamics). 
Most dynamical systems are non-linear and so one cannot obtain a linear system of the form $x_{t+1} = \mathcal{A} x_t$ to evolve in time from $t$ to $t+1$, with 
$\mathcal{A} \in \mathcal{R}^{N\times N}$.
While such models may work with locally linear approximations, many of the physics inherent in complex dynamical systems are
inherently complex and highly-nonlinear, e.g., flame zones in combustion systems, turbulent boundary layers, shock waves in supersonic airfoils, etc. 
In these regions linear approximations will fail, necessitating the development of non-linear models for dynamical systems. 
We will now describe different flavors of the deep Koopman model, use the notation $\mathcal{K}$ for the Koopman operator and $K$ for the Koopman matrix. 

\subsection{Koopman model}

In the classical Koopman dynamical model \cite{Koopman1931, Koopman1932}, the state vector $x_t$ is mapped on to a Hilbert space 
of possible measurements $y_t = g(x_t)$ of the state. The evolution of the system dynamics in time is linear in this Koopman invariant subspace,
and the infinite-dimensional Koopman operator $\mathcal{K}$ advances the system as:

\begin{equation}
\mathcal{K} g \left( x_t \right) = g \left( F \left( x_t \right) \right) =   g \left( x_{t+1} \right).
\end{equation}
The system is then projected back to the physical state vector space using an inverse function $g^{-1}$ \cite{Proctor2014, Kutz2016, Kaiser2017}:
\begin{equation}
g^{-1} \,\, \left( \mathcal{K} g \left( x_t \right) \right) = x_{t+1}.  
\end{equation}
For most real-world problems, however, evaluating $g$ and $g^{-1}$ functions accurately is not straightforward \cite{Mezic2004, Mezic2013, Arbabi17, Lusch2018, Salova19}.   
The Dynamic Mode Decomposition (DMD) algorithm \cite{Schmid2010, Kutz2016} can be used to obtain finite-dimensional representations of the Koopman operator using 
spatio-temporal coherent structures identified from a high-dimensional dynamical system. However, since it is based on a 
linearized analysis, it does not generally capture non-linear behaviors/transients accurately \cite{Lusch2018}.

\subsection{Deep Koopman model}

Recently, deep learning algorithms are used to accurately learn $g$ and $g^{-1}$ functions directly from data snapshots \cite{Takeishi2017, Lusch2018, Morton2018}.
Since neural networks are universal function approximators \cite{Hornik1989, Hornik1991}, they
are a natural choice to represent non-linear functions, particularly when the data is high-dimensional. 
An autoencoder \cite{AE} is used to represent the functions $g$ and $g^{-1}$, which are trained using data snapshots. 
The autoencoder consists of two neural networks, an encoder and a decoder.
The encoder learns the mapping function $g$ from $x_t$ to a latent embedding $z_t = g(x_t)$, represented as $g : \mathcal{R}^N \rightarrow \mathcal{R}^{M}$. 
The decoder learns the inverse mapping function $g^{-1}$, i.e., $\widehat{x_t} = g^{-1}(z_t)$, where $\widehat{x_t}$ is a reconstruction of $x_t$, and this
operation is mathematically represented as $g^{-1} : \mathcal{R}^M \rightarrow \mathcal{R}^{N}$. 
While we desire $\widehat{x_t} = x_t$ for accuracy, neural networks do have an inherent error which makes this impractical, particularly for 
high-dimensional data. Furthermore, note that $x \in \mathcal{R}^N$ and $z \in \mathcal{R}^M$, with $M \ll N$, and so the data dimension is reduced
by a few orders of magnitude. 
 
One can extend the system dynamics in time as follows. 
Consider time snapshots of the system $x_{1:T+1}$. One can construct the vectors: 
\begin{eqnarray}
X &=& \left[x_1, x_2, \cdots, x_T \right ] \nonumber \\
Y &=& \left[x_2, x_3, \cdots, x_{T+1} \right ]. 
\label{eq:XY}
\end{eqnarray} 
By feeding the snapshots into the encoder, one can obtain latent embeddings $Z = [z_1, z_2, \cdots, z_{T+1}]$, which can
be written as two separate vector of vectors:
\begin{eqnarray}
Z &=& \left[z_1, z_2, \cdots, z_T \right ] \nonumber \\
Z_{+1} &=& \left[z_2, z_3, \cdots, z_{T+1} \right ]. 
\label{eq:Z}
\end{eqnarray} 
A least squares fit was undertaken in \cite{Takeishi2017, Morton2018} to evaluate a $K$-matrix that can propagate the latent embeddings in time as
$K = Z_{+1} Z^{\dagger}$, where $Z^{\dagger}$ is the Moore-Penrose pseudoinverse.
The state embeddings are propagated in time as $Z_{+1}^{\mathrm{pred}} = K Z$ (note the superscript `pred' is used as this is the model's prediction
of the future latent embedding of the data).
Subsequently, $Z$ and $Z_{+1}^{\mathrm{pred}}$ are fed into the decoder--which learns the $g^{-1}$ function--to obtain the
reconstructed $\widehat{X}$ and the time advanced solution $\widehat{Y}$. 
The Deep Koopman model is trained to minimize the loss \cite{Takeishi2017, Morton2018}:
\begin{equation}
\mathcal{L} = \parallel X - \widehat{X} \parallel^2 + \parallel Y - \widehat{Y} \parallel^2.    
\end{equation}
Here, the first term is the typical autoencoder reconstruction loss \cite{AE} and the second term ensures that the time evolution of the dynamics is captured. 

In \cite{Takeishi2017}, $Z_{+1}^{\mathrm{pred}} = K \, Z$ was used, i.e, by advancing the dynamics by one time step for each snapshot. 
In \cite{Morton2018}, $Z_{+1}^{\mathrm{pred}}$ was evaluated by a recursive application of the $K$-matrix multiple times
to predict longer sequences of the system dynamics into the future:
\begin{equation}
Z_{+1}^{\mathrm{pred}} = \left[K \, z_1, K^2 \, z_1, \cdots, K^{T} \,z_1 \right].
\label{eqn:rec}
\end{equation}    
For the von Karman vortex shedding problem \cite{vonKarman, Roshko1955}, a 32-step sequence was demonstrated to be 
more accurate than the one-step version in \cite{Morton2018}.
In \cite{Lusch2018}, an embedding space loss: $\parallel Z_{+1}^{\mathrm{pred}} - Z_{+1} \parallel^2$ was also considered, with $Z_{+1}^{\mathrm{pred}}$ obtained from
Eqn. (\ref{eqn:rec}); in addition, they also used an $L_{\infty}$ loss to penalize the data point with the largest loss.

\subsection{Deep adversarial Koopman model}

\subsubsection{GAN loss}

In \cite{Balakrishnan20}, a Generative Adversarial Network (GAN) \cite{GAN} was also coupled to the Deep Koopman model by 
including an additional loss to train the ``generator'' networks.
It has been demonstrated in computer vision applications that coupling a GAN Discriminator with an autoencoder can improve the quality of samples
output from the autoencoder \cite{VAEGAN}. This is because the feature representations learned by the GAN discriminator also provide
additional learning signals for the primary neural networks, and this
improves the output quality \cite{VAEGAN}. 
In \cite{Balakrishnan20}, at every training iteration step a randomly sampled sequence of length $n_S$ was considered from the data corpus: $x_{t} \cdots x_{t+n_S}$.
Using this random sequence, the following vectors are constructed:
\begin{eqnarray}
X &=& \left[x_{t}, x_{t+1}, \cdots, x_{t+n_S-1} \right ] \nonumber \\
X_{+1} &=& \left[x_{t+1}, x_{t+2}, \cdots, x_{t+n_S} \right ] \nonumber \\
X_{+1}^{\mathrm{pred}} &=& \left[x_{t+1}^{\mathrm{pred}}, x_{t+2}^{\mathrm{pred}}, \cdots, x_{t+n_S}^{\mathrm{pred}} \right ].
\label{eq:gan1}
\end{eqnarray}
Here, $X_{+1}^{\mathrm{pred}}$ is the vector of the model's predictions for the states of the system at subsequent time steps.
The deep adversarial Koopman model \cite{Balakrishnan20} takes $x_t$ as input and outputs the sequence $X_{+1}^{\mathrm{pred}}$ using the Koopman 
dynamics recursively $n_S$ times. The GAN losses are then constructed using $X$, $X_{+1}$ and $X_{+1}^{\mathrm{pred}}$.
Two concatenated pairs were used: the ``real'' ($X,X_{+1}$) and ``fake'' ($X,X_{+1}^{\mathrm{pred}}$).
These pairs are fed into the GAN discriminator which outputs a single real value $D(\cdot)$ from which one can construct the GAN objective, following
the Wasserstein GAN \cite{WGANGP} approach due to its robustness against mode collapse: 
\begin{equation}
L^{\mathrm{GAN \, objective}} = \mathop{\mathbb{E}}_{x\in (X,X_{+1})} \left[ D(x) \right ] - \mathop{\mathbb{E}}_{\widetilde{x}\in (X,X_{+1}^{\mathrm{pred}})} \left[ D(\widetilde{x}) \right ]. 
\end{equation}
This additional loss term was used in the adversarial Koopman model and was found to robustify the overall predictions. In \cite{Balakrishnan20}, this
approach was demonstrated to work on the Gray-Scott reaction-diffusion problem \cite{Pearson1993} which is a widely used model for
investigating the Turing instability \cite{Turing1952}.

\subsubsection{Auxiliary network}
To obtain the Koopman matrix $K$, \cite{Morton2018} used a simple least squares approach.
In \cite{Lusch2018}, an auxiliary neural network was used to obtain the eigenvalues of $K$, 
from which the $K$ matrix was constructed using Jordan block structure. 
In \cite{Balakrishnan20}, the auxiliary neural network was used to output the full $K$ matrix ($M \times M$).
In this study, we will consider a similar approach, but we will in addition also consider tridiagonal Koopman matrices and 
demonstrate good results with this assumption.

\subsubsection{Residual Koopman}
\label{sec:resK}
\cite{Balakrishnan20} also used a residual Koopman approach where the system dynamics instead of being represented as $z_{t+1} = \mathcal{K} z_{t} = K z_{t}$, 
was represented as $\mathcal{K} z_{t} = z_{t+1} = z_{t} + K z_{t}$ (note that $\mathcal{K}$ is the Koopman operator, whereas $K$ is the corresponding Koopman matrix). 
Here, the Koopman model learns the residual change required to advance the system in time in the latent embedding space. 
With this change, the recursive prediction of the future states of the system in the embedding space is:
\begin{equation}
Z_{+1}^{\mathrm{pred}} = \left[\mathcal{K} \, z_t, \mathcal{K}^2 \, z_t, \cdots, \mathcal{K}^{n_S} \,z_t \right],
\end{equation}
where $\mathcal{K}^{j}$ represents the application of the Koopman opertaor $\mathcal{K}$ $j$ times, and $n_S$ is the desired sequence length.

\subsection{Stochastic adversarial Koopman model}

\subsubsection{Stochastic embedding}

\begin{figure}[h]
\centering
\includegraphics[width=12cm]{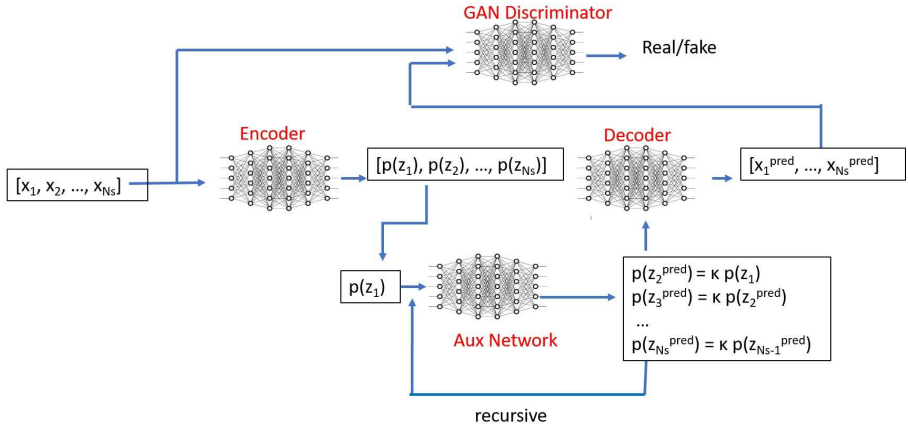}  
\caption{Schematic of the Stochastic Adversarial Koopman Model showing the different neural networks and the connections.}
\label{fig:koopmanschematic}
\end{figure}

In the new stochastic adversarial Koopman (SAK) approach, we model the latent embedding $z_t$ as a Gaussian random variable. 
Thus, $z_t$ involves two components $\mu^z_t$ and $\sigma^z_t$, each of which $\in \mathcal{R}^M$. The auxiliary network takes
$\mu^z_t$ and $\sigma^z_t$ as input and outputs two Koopman matrices $K_{\mu} (\mu^z_t, \sigma^z_t)$ and 
$K_{\sigma} (\mu^z_t, \sigma^z_t)$
which are used to construct the distribution for the next time step using the residual approach described in Section $\ref{sec:resK}$:
\begin{eqnarray}
\mu^z_{t+1} = \mu^z_{t} + K_{\mu} \mu^z_{t}; \,\,\, ln \, \sigma^z_{t+1} = ln \, \sigma^z_{t} + K_{\sigma} ln \, \sigma^z_{t}.
\end{eqnarray}
Note the use of $ln \, \sigma^z_{t}$ and not $\sigma^z_t$ as using the latter will not guarenteee positive standard deviations after the matrix multiplication
operation \footnote[1]{$ln$ is the natural logarithm to the base $e$}. Essentially, this translates as a power law in $\sigma^z$, with the Koopman 
matrix $K_{\sigma}$ representing the incremental change in the exponent to proceed from one time step to the next in $\sigma^z$. 
Once the distributions for the future states are obtained, we sample $z_{t+1} = \mathcal{N}(\mu^z_{t+1},\sigma^z_{t+1})$, which
is then passed on to the decoder. 
A schematic of the SAK model is shown in Fig. \ref{fig:koopmanschematic}. 
The use of distributions for data in the latent embedding space is powerful as one can 
now handle uncertainties inherent in complex dyanmical systems such as observed in chaos, bifurcation and fluid turbulence problems where slight perturbations can result in completely different
dynamics of the system. Moreover, learning the data distribution helps as it can also be used as a generative model to
obtain a family of trajectories of a given dynamical system as expectations defined by the learned statistics.

\subsubsection{Loss terms}
The encoder and decoder are represented as $g(\cdot)$ and $g^{-1}(\cdot)$, respectively. 
We use the mean squared error (MSE) and the maximum mean discrepancy (MMD) \cite{WAE} to construct different loss terms:
(1) reconstruction loss $L^{\mathrm{recon}}$, (2) prediction loss $L^{\mathrm{pred}}$,
(3) code loss $L^{\mathrm{code}}$, (4) gradient loss $L^{\mathrm{grad}}$, (5) $L_2$ regularization loss $L^{\mathrm{reg}}$,
(6) GAN loss $L^{\mathrm{GAN}}$, and (7) discriminator loss $L^{\mathrm{disc}}$. These different losses are summarized below:

\begin{equation}
L^{\mathrm{recon}} = \parallel x_t - g^{-1} g \left( x_t \right)  \parallel_{\textrm{MSE}}
\end{equation}

\begin{equation}
L^{\mathrm{pred}} = \frac{1}{n_S} \sum_{m=1}^{n_S} \parallel x_{t+m} - g^{-1} \left( \mathcal{K}^m g \left( x_t \right) \right) \parallel_{\textrm{MSE}}
\end{equation}

\begin{equation}
L^{\mathrm{code}} = \mathrm{MMD} [ g \left( x_{t+m} \right), \mathcal{K}^m g \left( x_t \right) ]
\end{equation}

\begin{eqnarray}
L^{\mathrm{grad}}_j &=& \frac{1}{n_S} \sum_{m=1}^{n_S} \parallel \nabla_j \left[ x_{t+m} - g^{-1} \left( \mathcal{K}^m g \left( x_t \right) \right) \right]  \parallel_{\textrm{MSE}}, \textrm{$j$=1, 2, 4} \nonumber \\
L^{\mathrm{grad}} &=& \lambda_1 L^{\mathrm{grad}}_1 + \lambda_2 L^{\mathrm{grad}}_2 + \lambda_4 L^{\mathrm{grad}}_4
\end{eqnarray}

\begin{equation}
L^{\mathrm{reg}} = \lambda_{\mathrm{reg}} \sum w_i^2 
\end{equation}

\begin{equation}
L^{\mathrm{GAN}} = \mathop{\mathbb{E}}_{\widetilde{x}\in (X,X_{+1}^{\mathrm{pred}})} \left[ D(\widetilde{x}) \right ]
\end{equation}

\begin{equation}
L^{\mathrm{disc}} = \mathop{\mathbb{E}}_{\widetilde{x}\in (X,X_{+1}^{\mathrm{pred}})} \left[ D(\widetilde{x}) \right ] - \mathop{\mathbb{E}}_{x\in (X,X_{+1})} \left[ D(x) \right ]
\end{equation}

\noindent Note that the losses $L^{\mathrm{recon}}$, $L^{\mathrm{pred}}$,
and $L^{\mathrm{code}}$ were also considered in \cite{Lusch2018, Balakrishnan20}, but MSE was considered for all these terms.
In this study, since the latent embedding is stochastic, we use the MMD loss \cite{WAE}: 
\begin{eqnarray}
L^{\mathrm{code}} = \frac{1}{n_S (n_S-1)} \sum_{l, j, l \ne j} f \left(z_l, z_j \right) + \frac{1}{n_S (n_S-1)} \sum_{l, j, l \ne j} f \left(z_l^{\mathrm{pred}}, z_j^{\mathrm{pred}} \right) - \frac{2}{n_S^2} \sum_{l, j} f \left(z_l, z_j^{\mathrm{pred}} \right), 
\end{eqnarray}
where $z_l$ is sampled from $\mathcal{N}(g\left( x_{t+l} \right))$ and $z_l^{\mathrm{pred}}$ from $\mathcal{N}( \mathcal{K}^l g\left( x_{t} \right))$. Note that we essentially have samples from 
two distributions $g\left( x_{t+l} \right)$ and  $\mathcal{K}^l g\left( x_{t} \right)$ that need to be synchronized and so the Maximum Mean Discrepancy (MMD) \cite{WAE} is one such
cost function. Other optimal transport loss such as Sinkhorn divergence \cite{Cuturi2013} can also be considered in the future.   
Note that the function $g$ takes $x_t$ as input and outputs $\mu^z_t$ and $\sigma^z_t$. And $g^{-1}$ takes $z_t$ (or $z_{t+1}$) as input to output $x_t$ (or $x_{t+1}$).
$f$ is the inverse multiquadratics kernel $f(x,y) = C/(C+\parallel x-y \parallel_2^2)$ with $C$ being a constant \cite{WAE}.
In the above equations, $\mathcal{K}^m$ denotes the application of the Koopman operator $m$ times.
Gradient losses are also used to improve the overall quality \cite{Mathieu2015} of the output and $\nabla_1$, $\nabla_2$ and $\nabla_4$ are the 
first, second and fourth derivatives, respectively. We use only the first derivative for the gradient loss for all the problems in this paper, except for the 
Kuramoto-Sivashinsky (KS) test case for which all the three gradient terms are used. 

A total of four neural networks are used: the encoder, decoder, auxiliary network and the discriminator and their architectures are summarized in Appendix B.
The encoder, decoder and the auxiliary network---which we can refer to as the ``generator'' in GAN parlance---are trained jointly using the loss function:
\begin{equation}
L^{\mathrm{total}} = L^{\mathrm{recon}} + L^{\mathrm{pred}} + \lambda_{\mathrm{code}} L^{\mathrm{code}} + \lambda_{\mathrm{grad}} L^{\mathrm{grad}} + \lambda_{\mathrm{reg}} L^{\mathrm{reg}} + \lambda_{\mathrm{GAN}} L^{\mathrm{GAN}}, 
\end{equation}
and the GAN discriminator is trained using $L^{\mathrm{disc}}$. 
We use $\lambda_{\mathrm{code}}$ = 100, $\lambda_{\mathrm{grad}}$ = 1 and $\lambda_{\mathrm{reg}}$ = 10$^{-3}$ for all the cases considered in this paper.
$\lambda_{\mathrm{GAN}}$ varies for each test case and is obtained from experimentation; it is typically in the range 0--0.1 for best results. 
The gradient loss terms are summarized in Appendix B of \cite{Balakrishnan20} and not shown here for brevity.
The discriminator is trained by minimizing $L^{\mathrm{disc}}$, along with an additional gradient penalty loss term similar to WGAN-GP \cite{WGANGP}. 
We alternate between training the generator one step followed by the discriminator for one step, at each training iteration.
At test time, the discriminator is not used.


\section{Experiments}

We will now demonstrate the robustness of the Stochastic Adversarial Koopman (SAK) model on a variety of engineering test problems.

\subsection{Test cases}

We consider five different problems to test the performance of the SAK model. These test problems involve 
a series of partial differential equations (PDEs) and finite difference methods are used to 
solve the governing equations. This creates the data corpus on which we will train the SAK model.
The test problems involve both one- and two-dimensional in space, and so the convolutional filters in the SAK model are
varied as appropriate.      
The test problems we consider are (1) the Kuramoto-Sivashinsky (KS) equation \cite{Kuramoto1978, Sivashinsky1977} that is
widely used to model chaos; (2) von Karman vortex shedding behind a cylinder \cite{vonKarman}; (3) Flame ball-vortex interaction \cite{Roussel05}; (4) FitzHugh-Nagumo model;
and (5) Doyle-Fuller-Newman model for Lithium ion batteries \cite{Dualfoil1, Dualfoil2}. 
These equations are first solved using well established finite difference schemes to obtain the data corpus, which is 
used to train the SAK models in this paper. Note that the finite difference methods required to 
solve these equations require much finer time step $\Delta t$ and so the $\Delta t$ used to obtain the data corpus is
different from the $\Delta t$ used in the Koopman analysis.
These five test cases, the CFD methods to solve them, and the $\Delta t$ used in the Koopman analysis are summarized in Appendix A.


\subsection{Predictions of the SAK model}
\label{sec:dakm}

\subsubsection{KS data}

\begin{figure}[h]
\centering
(a) \includegraphics[width=12cm]{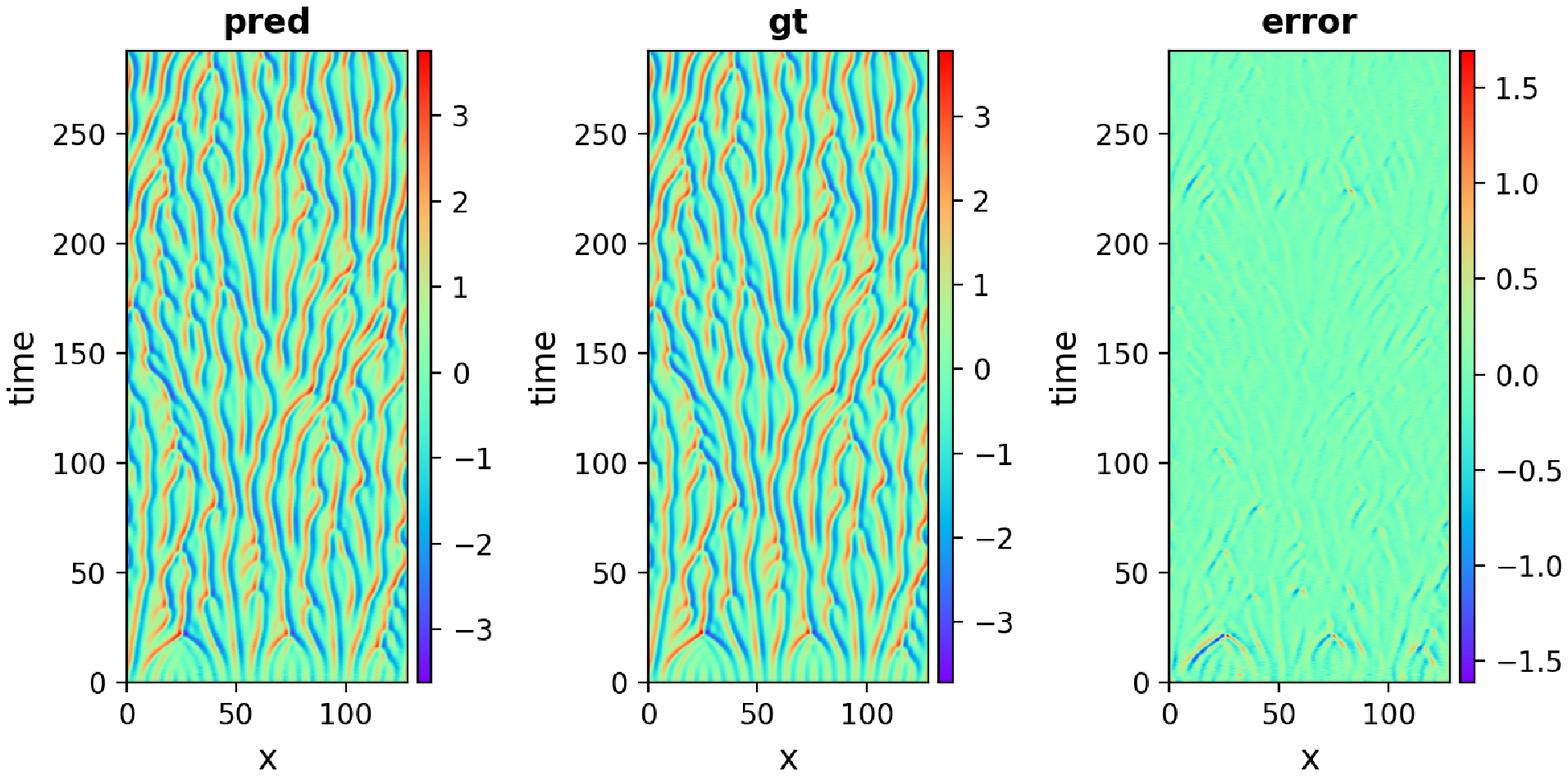} \\  
(b) \includegraphics[width=7.5cm]{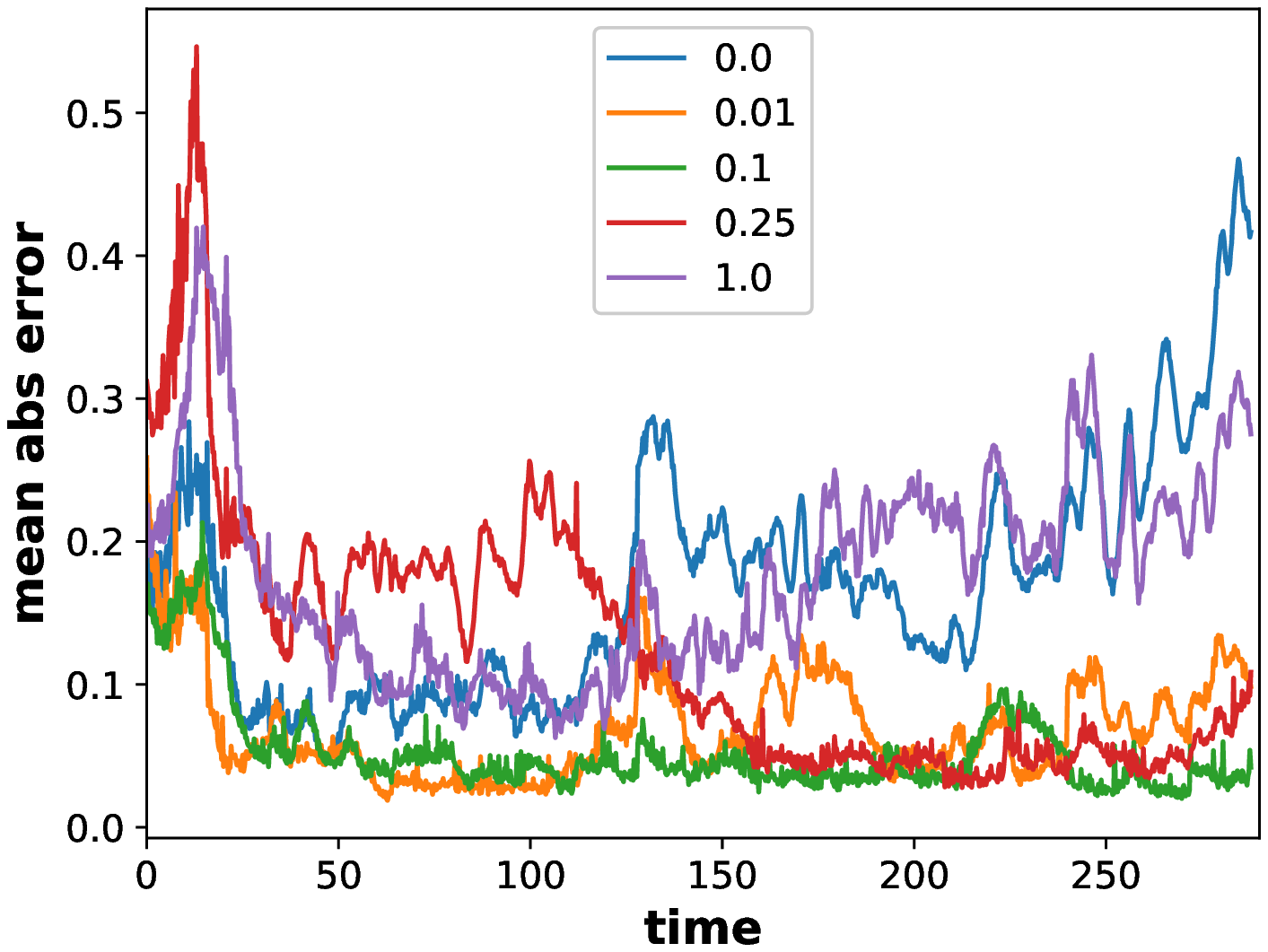}
\caption{Predictions of $u(x,t)$ for the Kuramoto-Sivashinsky equation with the SAK model: (a) comparison of the model predictions (``pred''), ground truth (``gt''), and the error (difference) between gt and pred (``error''); (b) mean absolute error for different $\lambda_{\mathrm{GAN}}$ values shown in the legend.}
\label{fig:ks1}
\end{figure}

The SAK model is trained on the Kuramoto-Sivashinsky (KS) data corpus using $n_S$ = 64 time-step sequences for 100k iterations, where at each iteration a
contiguous sequence of $n_S$ snapshots are randomly sampled and used to train the network. The hyperparameter $\lambda_{\mathrm{GAN}}$ controls the trade-off
between the GAN loss vis-\'a-vis other loss terms. To better understand the role of the GAN losses, we repeat the training for
multiple $\lambda_{\mathbf{GAN}}$ values: 0, 0.01, 0.1, 0.25, 1.0, and compare the predictions at test time.
Specifically, at test time, we make predictions starting from time = 0 to time = 288 (i.e., for a total of 
1152 time steps, since $\Delta t$ = 0.25). This corresponds to 18 cycles, where each cycle is a sequence of 64 time steps with the last prediction from each 
cycle used as input to the next cycle. The model predictions (``pred'') and ground truth (``gt'') of the chaotic patterns at different ($x,t$) are shown
in Fig. \ref{fig:ks1} (a) for $\lambda_{\mathrm{GAN}}$ = 0.1, and demonstrate good
agreement. The error is near zero for the most part, except at early times since the patterns follow a transient trajectory at early times which is different from
the later time natural trajectories of the system. Thus, the SAK model has learned to replicate the training data. Note that this problem is not periodic and therefore it was found to 
not make accurate predictions of future states of the system, unlike the studies of \cite{Lusch2018, Morton2018} where a similar model was applied to
periodic problems such as the von Karman vortex shedding problem \cite{vonKarman}. However, our model is still able to capture the chaotic 
dynamics of the system and is able to reproduce the training data well.   

The mean absolute error at every time step of the prediction is shown in Fig. \ref{fig:ks1} (b) for different $\lambda_{\mathrm{GAN}}$ values.
As evident, $\lambda_{\mathrm{GAN}}$ = 0.1 has the lowest prediction errors and the errors are higher for $\lambda_{\mathrm{GAN}}$ values on either side of the 
optimal value. This demonstrates that the amount of GAN loss required 
must be optimized for best performance, and the errors are strongly dependent on the choice of the trade-off parameter $\lambda_{\mathrm{GAN}}$.
Using GAN loss robustifies the predictions, provided that the right balance between the GAN and other losses are used.    
Our experience suggests that the optimal value of $\lambda_{\mathrm{GAN}}$ varies for different problems and its estimation requires experimentation.


\subsubsection{von Karman vortex shedding data}

\begin{figure}[h]
\centering
(a) \includegraphics[width=13cm]{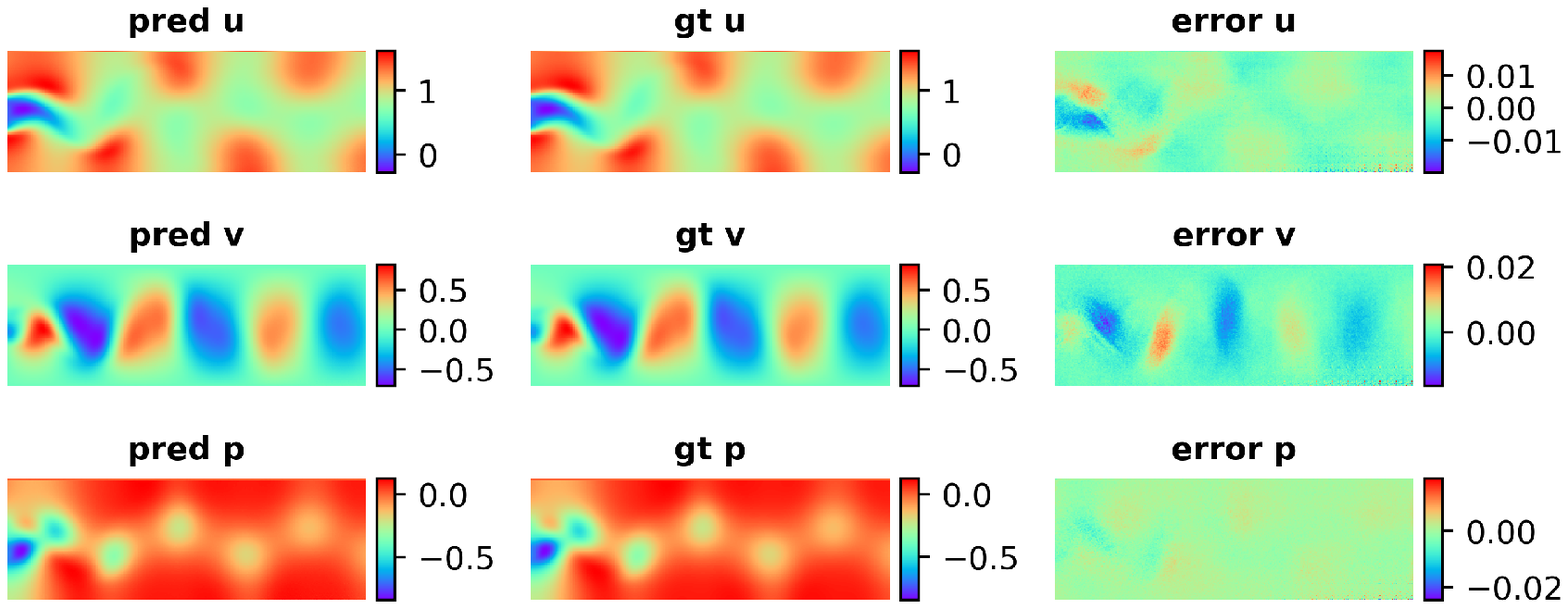} \\
(b) \includegraphics[width=13cm]{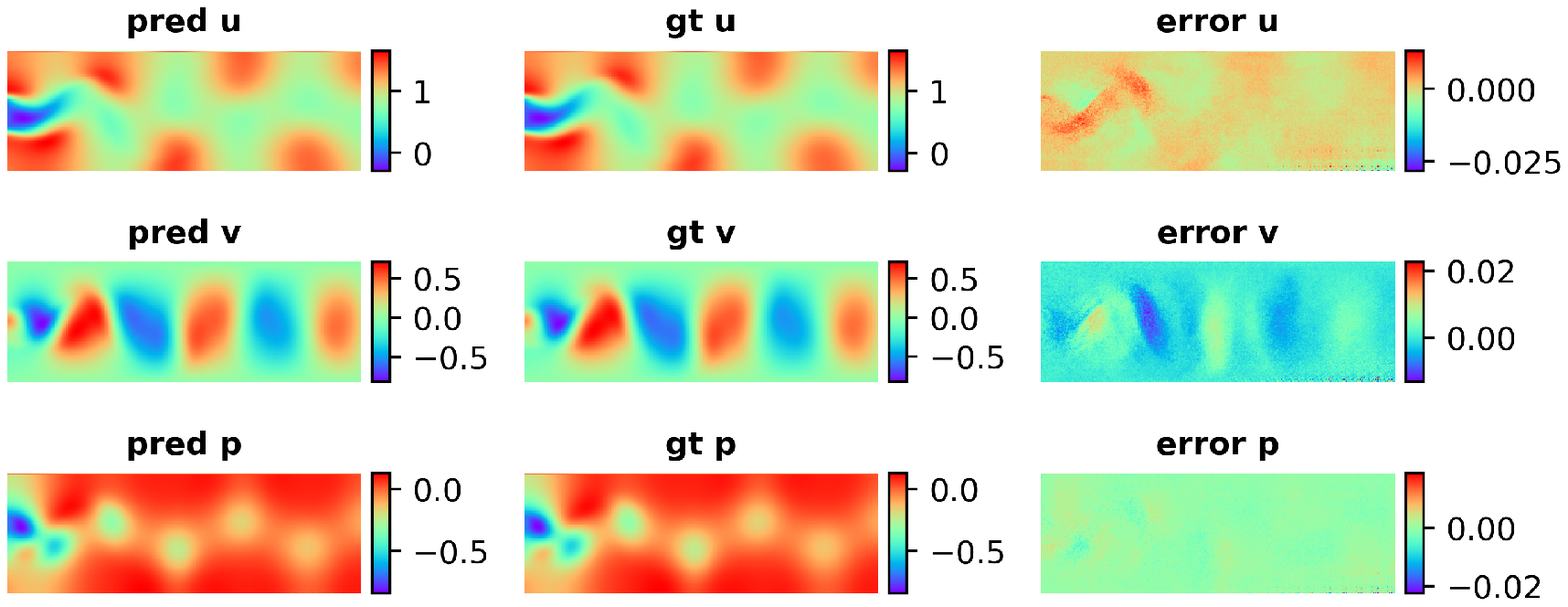}
\caption{Predictions of $u$, $v$ and $p$ for the von Karman vortex shedding problem behind a cylinder with the SAK model at two different times half a period apart. The model predictions are denoted as ``pred,'' ground truth as ``gt,'' the  and the error as ``error.''}
\label{fig:karman1}
\end{figure}

Next, the SAK model is trained on the von Karman vortex shedding data. 
For this problem, we use 2D kernels in the encoder, decoder and GAN discriminator networks. 
In addition, we also vary the sequence length $n_S$ from 16 to 32 as the training progresses. 
At early times, the sequence length is shorter so that the network learns the problem dynamics faster. As training progresses, we gradually increase $n_S$ so that 
the network is now tasked to learn slightly longer time sequences. Specifically, $n_S$ = 16 at the start of the training, and is gradually increased (i.e., by 5\%) every 20k iterations 
following the curriculum:
\begin{equation}
n_S \leftarrow \mathrm{min} \left[ \mathrm{round} \left(1.05 \times n_S \right) +1, 32 \right]. 
\end{equation}
The value of $n_S$ is increased gradually until $n_S$ = 32, after which it is held fixed due to GPU memory restrictions \footnote[2]{We used an Nvidia Geforce RTX 2080 Ti GPU with 11GB memory; the largest value of $n_S$ depends on the data dimension which varies for the different problems}. 
The hyperparameter $\lambda_{\mathrm{GAN}}$ = 0.01 was found to be the optimal value (i.e. least error) for this problem after some experimentation.
At test time, we make predictions
for 10 cycles, where each cycle is a sequence of 32 time steps with the last prediction from each 
cycle used as input to the next cycle. The model predictions (``pred'') and ground truth (``gt'') of the velocity components $u$, $v$ and pressure $p$ at 
two different time instants half a period apart are shown in Fig. \ref{fig:karman1}. As evident, the model predictions are 
very similar to the ground truth values and the error magnitudes are very low. 

\begin{figure}[h]
\centering
\includegraphics[width=8cm]{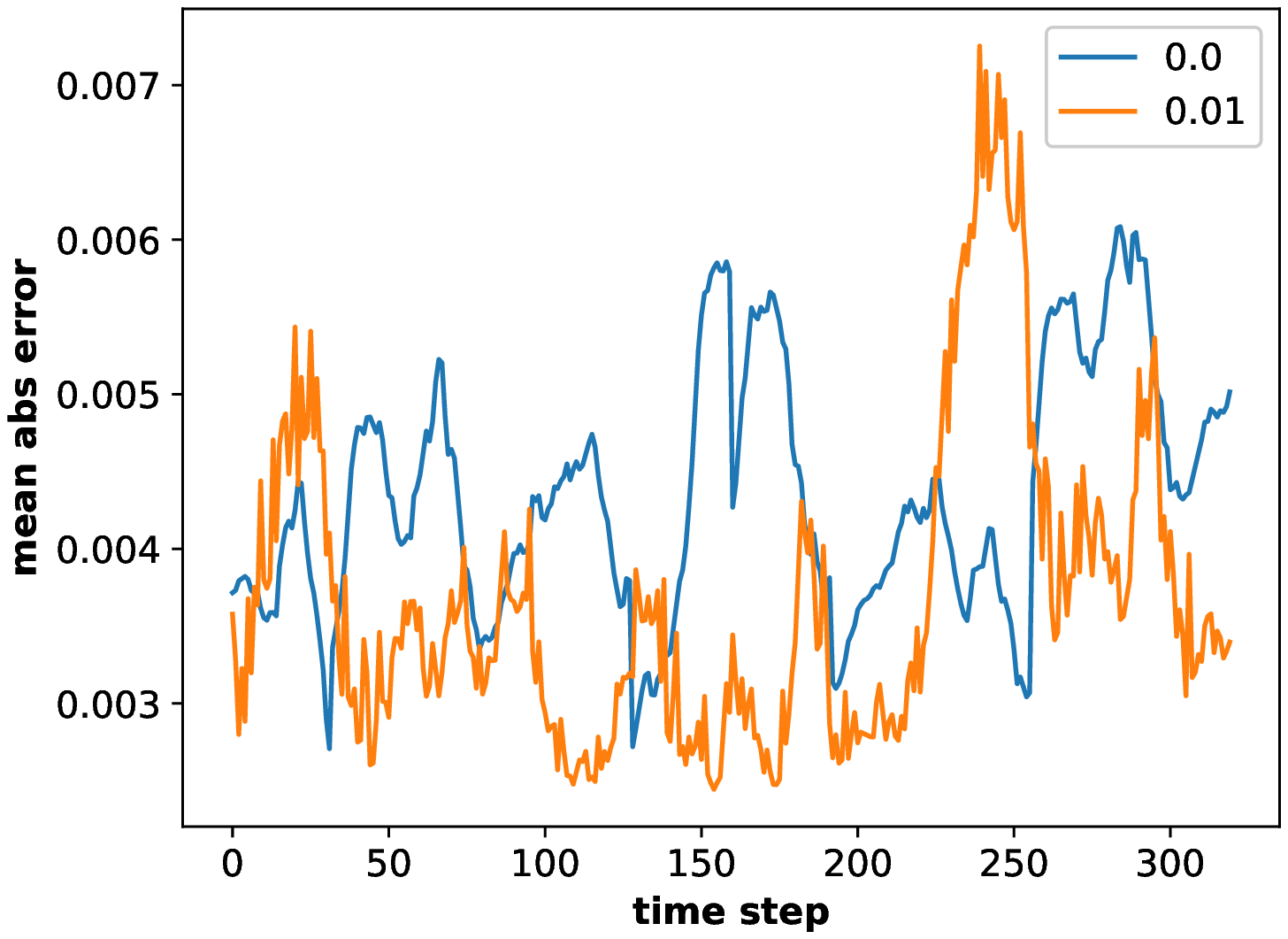} 
\caption{Prediction error for each time snapshot for the von Karman vortex problem with $\lambda_{\mathrm{GAN}}$ = 0.0 and 0.01.}
\label{fig:karman2}
\end{figure}

To better understand the role of the GAN loss, we repeat the training from scratch with the GAN loss turned off, i.e., $\lambda_{\mathrm{GAN}}$ = 0.0.
In Fig. \ref{fig:karman2}, we compare the test time prediction errors with and without the GAN loss term used during training.    
As evident, $\lambda_{\mathrm{GAN}}$ = 0.01 has lower prediction errors compared to $\lambda_{\mathrm{GAN}}$ = 0.0 for most time steps,
although we observe one small time interval where using the GAN loss during training results in higher errors. 
Since at most time instants the errors are lower with $\lambda_{\mathrm{GAN}}$ = 0.01, we conclude that using the GAN loss
at training helps in lowering the test time prediction errors for most time instants, thereby having an overall beneficial effect. 
As aforementioned, this benefit in lowering test time errors is however sensitive to the choice of the trade-off
parameter $\lambda_{\mathrm{GAN}}$, which ahs to be calibrated for every problem by experimentation.

\subsection{Tridiagonal Koopman matrix}

\begin{figure}[h]
\centering
\includegraphics[width=10cm]{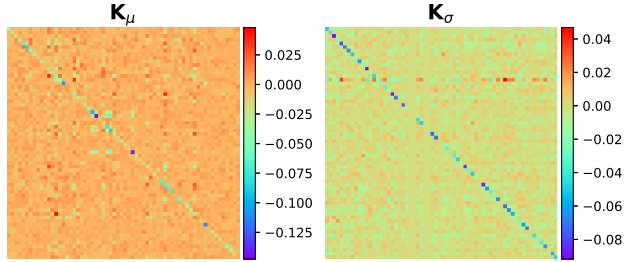}
\caption{$K_{\mu}$ and $K_{\sigma}$ for the von Karman vortex shedding problem at the same time instant as Fig. \ref{fig:karman1} (a).}
\label{fig:kmat}
\end{figure}

The structure of the Koopman matrices is of particular interest. 
We have hitherto used full Koopman matrices $K_{\mu}$ and $K_{\sigma}$.
In \cite{Lusch2018} a Jordan block structure was used for the Koopman matrix, whereas in \cite{Salova19}, block diagonal structure was used.
A tridiagonal Koopman matrix, with the entries above and below the diagonal being asymmetric, was considered in \cite{Pan2020} and also produced good results. 
To better understand the magnitudes of the Koopman matrices, we present $K_{\mu}$ and $K_{\sigma}$ in Fig. \ref{fig:kmat}
at the same time instant as Fig. \ref{fig:karman1} (a).
As evident, the entries along the diagonal are preponderant for both $K_{\mu}$ and $K_{\sigma}$. Although a few off-diagonal 
entries also have relatively large magnitude values, no structure is evident other than the diagonal preponderance.

Inspired by \cite{Pan2020}, we now focus on tridiagonal Koopman matrices, albeit without the asymmetric constraints of \cite{Pan2020}, and compare the model predictions
with that presented in Fig. \ref{fig:karman1} (which used full Koopman matrices).
The model is trained on the von Karman vortex data again from scratch, but with $K_{\mu}$ and $K_{\sigma}$ being tridiagonal matrices. 
The output layer of the auxiliary network is modified to output $M$ + 2$\times$($M$-1) = 3$M$-2 entries instead of $M^2$ for each 
of the matrices $K_{\mu}$ and $K_{\sigma}$. Thus, we have reduced the dimension of the Koopman matrices from $O(M^2)$ to $O(M)$.
(Note that $M$ = 64 is used throughout this study.)

\subsubsection{von Karman vortex shedding data}

\begin{figure}[h]
\centering
(a) \includegraphics[width=13cm]{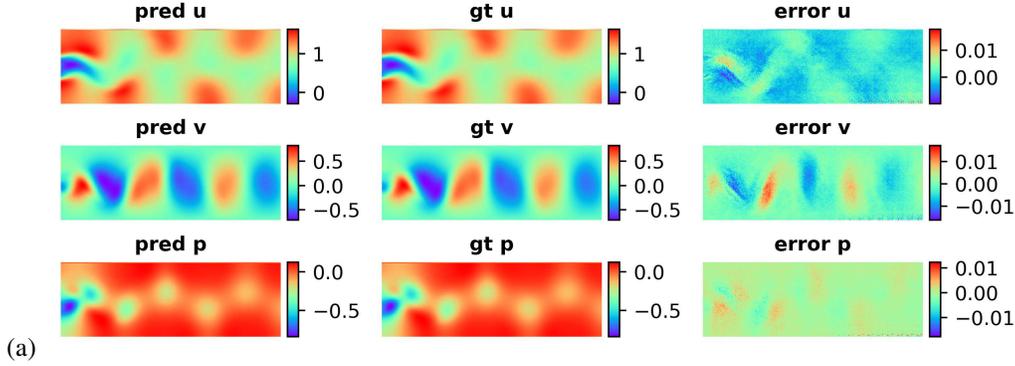} \\
(b) \includegraphics[width=8cm]{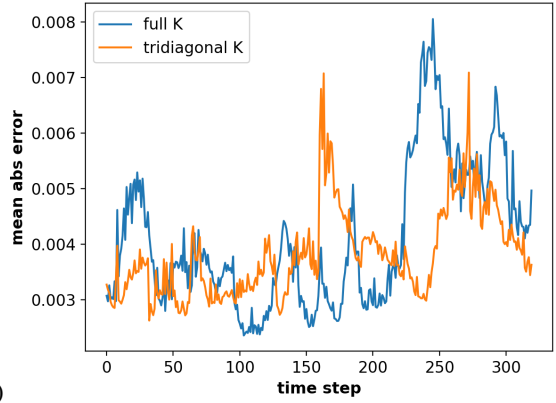}
\caption{Predictions using tridiagonal Koopman matrices in the SAK model for the von Karman vortex shedding problem: (a) flow-field at the same time instant as Fig. \ref{fig:karman1} (a), the model predictions are denoted as ``pred,'' ground truth as ``gt,'' the  and the error as ``error''; (b) mean absolute error for every time snapshot w.r.t. the ground truth for the full and tridiagonal Koopman matrices.}
\label{fig:karman3}
\end{figure}

Using the tridiagonal Koopman approximation, we re-train the SAK model on the von Karman vortex shedding problem. The model prediction results at the same time
instant as Fig. \ref{fig:karman1} (a) are shown in Fig. \ref{fig:karman3} (a), and as apparent the reduced Koopman matrices suffice for learning the 
dynamics in the Koopman space. We also compare the mean absolute errors at each time step for the predictions using the full Koopman matrices vis-\'a-vis the
reduced tridiagonal Koopman formulations in Fig. \ref{fig:karman3} (b). The errors are nearly similar in magnitude for both formulations despite having
one order of magnitude difference in the number of non-zero values in  $K_{\mu}$ and $K_{\sigma}$. We observe one small zone
of high error for each of the two formulations, albeit at different time instants. A reduced tridiagonal Koopman matrix 
may thus suffice for many problems. The tridiagonal Koopman demonstrated here can be seen as an alternative to the 
Jordan/diagonal block structures for $K$ used in \cite{Lusch2018, Salova19}. For the remainder of this paper, we will use the reduced tridiagonal Koopman formulation, unless
otherwise stated.

\subsubsection{Flame ball-vortex interaction}

\begin{figure}[h]
\centering
(a) \includegraphics[width=12cm]{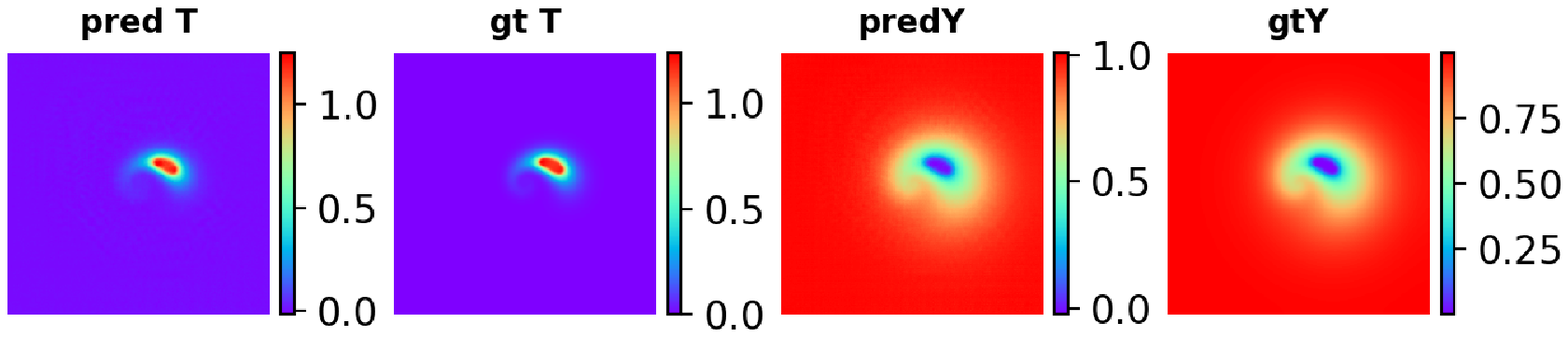} \\
(b) \includegraphics[width=12cm]{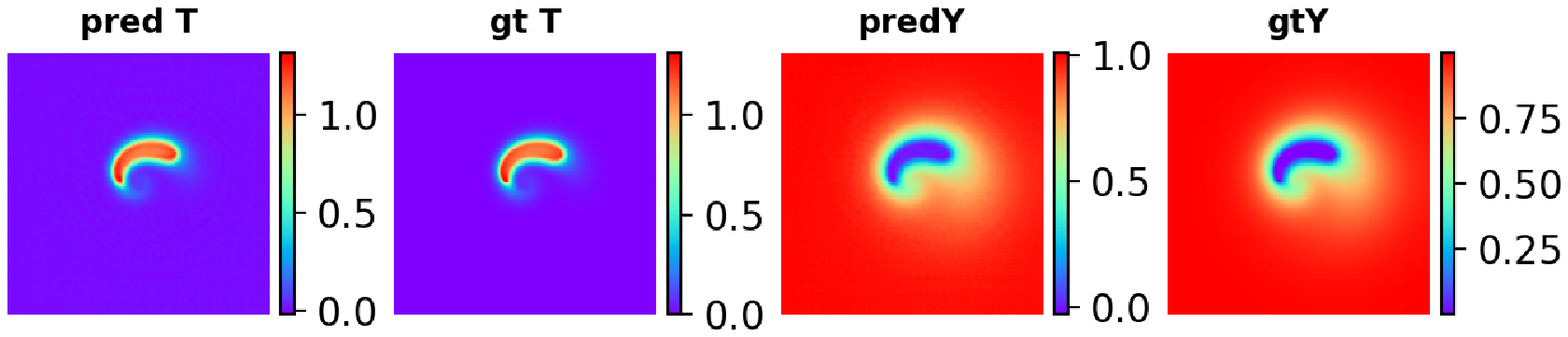} \\
(c) \includegraphics[width=12cm]{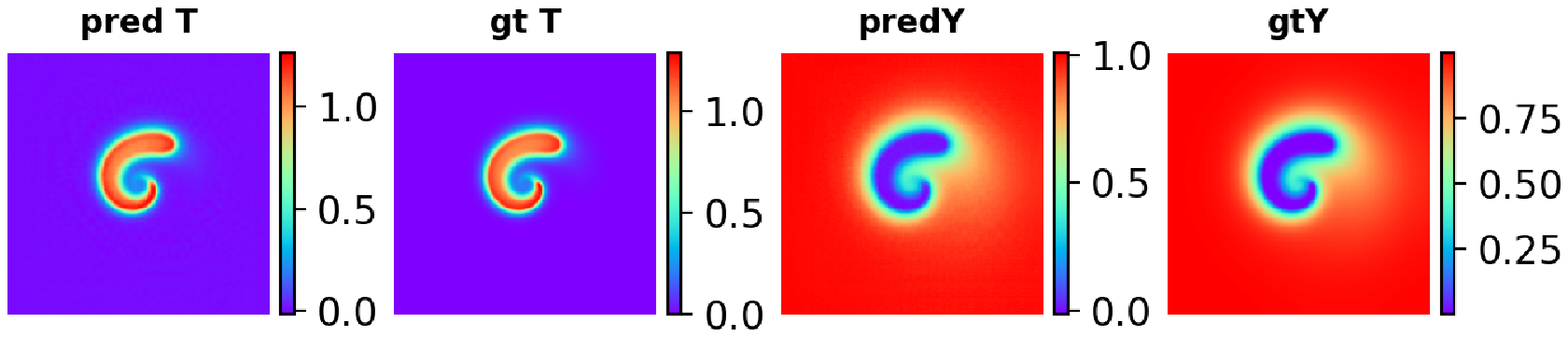} \\
(d) \includegraphics[width=12cm]{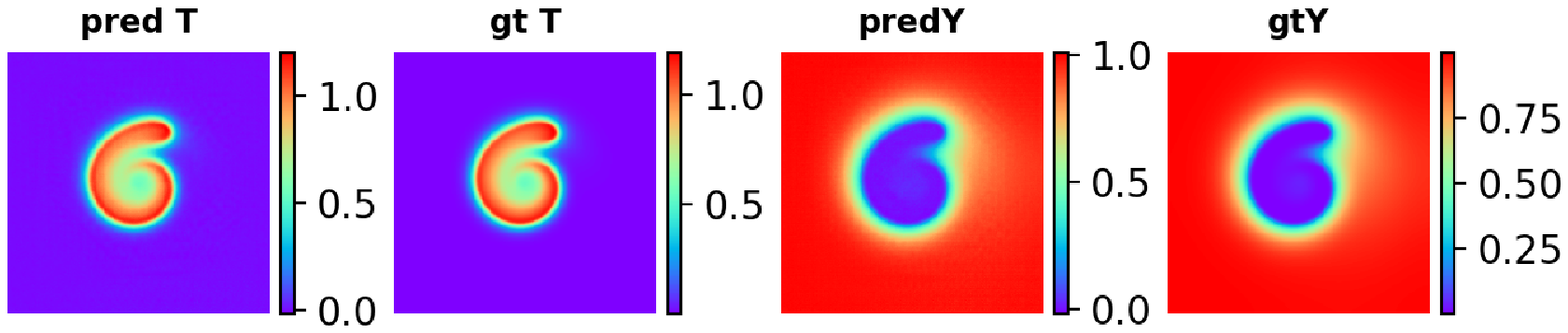} 
\caption{Predictions of $T$ and $Y$ for the flame ball-vortex interaction problem with the SAK model at time step number: (a) 22, (b) 53, (c) 78 and (d) 112. The titles ``pred'' and ``gt'' denote the SAK model prediction and the ground truth, respectively.}
\label{fig:fv1}
\end{figure}

We use the tridiagonal Koopman matrix in the SAK model to train the flame ball-vortex interaction data corpus. $\lambda_{\mathrm{GAN}}$ = 0.01 and 300k iterations are
used for training the model. The model predictions at test time, compared with the ground truth and the prediction error are presented in Fig. \ref{fig:fv1}.
The SAK model is able to accurately capture the shape of the flame ball as the vortex distorts it counter-clockwise. At late time
the surface area of the flame increases, which further consumes more of the pre-mixed fuel-oxidizer mixture. 
The errors (not shown) are very small compared to the magnitudes of the 
variables, demonstrating the efficacy of the SAK model. This problem of flame-vortex interaction is a fundamental problem in combustion engines and 
Koopman family of models can be used to speed-up the design and development of the next generation of combustion engines.

\subsubsection{FitzHugh-Nagumo model}

\begin{figure}[h]
\centering
(a) \includegraphics[width=8.5cm]{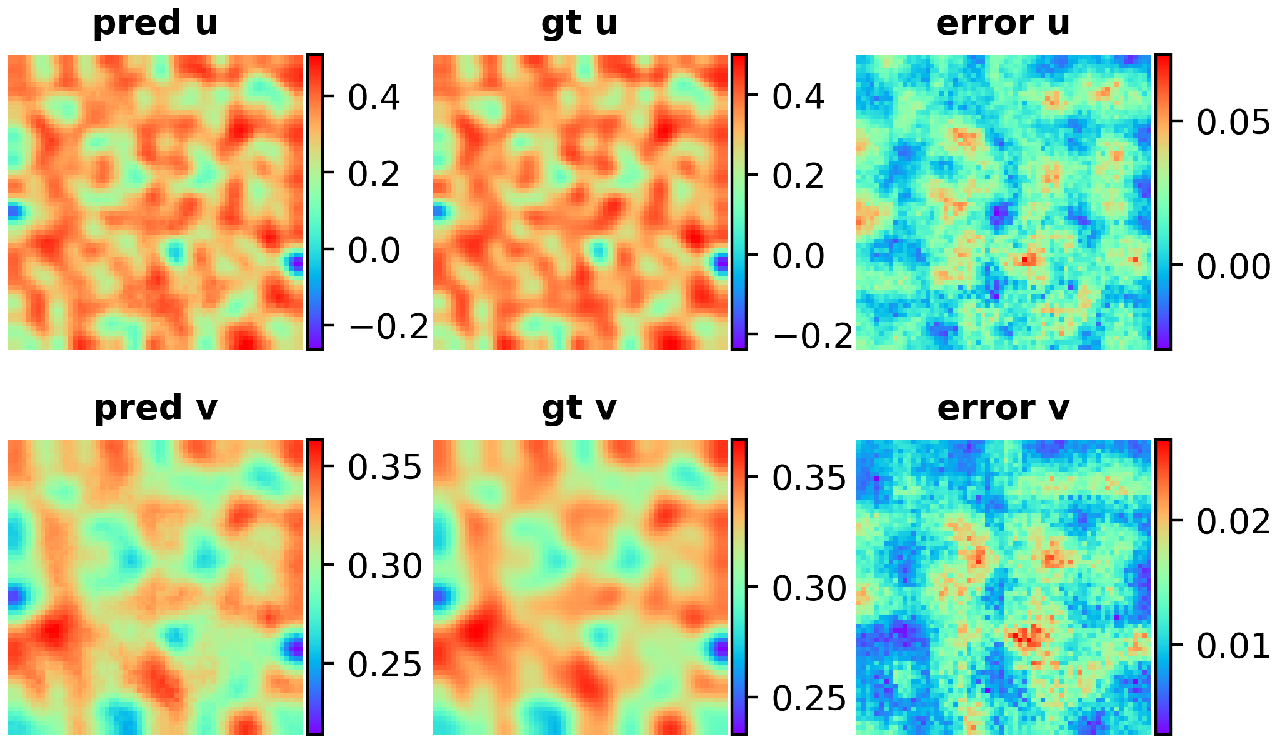} \\  
(b) \includegraphics[width=8.5cm]{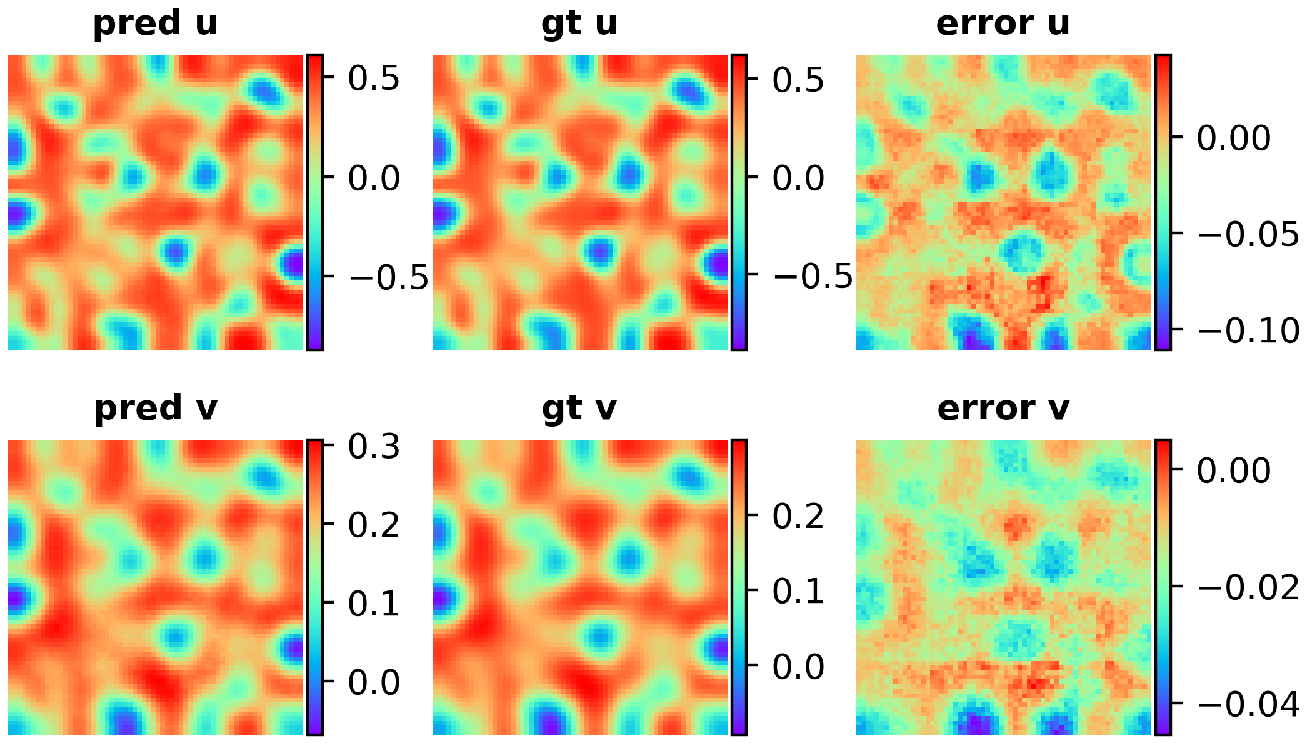} 
\caption{Predictions of $u$ and $v$ in the FitzHugh-Nagumo model using the SAK model at time step number: (a) 20 and (b) 100. The titles ``pred'' and ``gt'' denote the SAK model prediction and the ground truth, respectively, and ``error'' is the difference between them.}
\label{fig:fn1}
\end{figure}

The SAK model using the tridiagonal Koopman matrix is trained on the FitzHugh-Nagumo model data corpus. 
Note that this problem is hard to model compared to the other problems considered above due to 
the inherent stochasticity of the FitzHugh-Nagumo model.  
We use 300k iterations and considered different values of $\lambda_{\mathrm{GAN}}$ in the range 0-0.01, and found $\lambda_{\mathrm{GAN}}$ = 0.0 to 
yield the best results (comparison of results with different $\lambda_{\mathrm{GAN}}$ values not shown for brevity). 
Thus, the effect of using a GAN discriminator to improve the Koopman model predictions varies for
different problems. The predictions of $u$ and $v$ using the model (``pred''), compared with the
ground truth (``gt'') and the prediction error (``error'') are presented in Fig. \ref{fig:fn1}.
The SAK model is able to accurately capture the patterns in the systems.  
At early times (Fig. \ref{fig:fn1} a), the patterns are more sporadic as we start with a random initialization. 
However, at later times (Fig. \ref{fig:fn1} b), the patterns are more organized.
The errors in the prediction are mostly localized, and this problem demonstrates 
the robustness of the SAK model for such chaotic dynamical systems.

\subsection{Conditional SAK model on the DFN data}

The last test case on which we will demonstrate the SAK model is the DFN battery data set. As aforementioned, here we will experiment a conditional 
setting where the applied current density $I_{app}$ (in A/m$^2$) is fed as input to the auxiliary neural network, so that the Koopman matrices
$K_{\mu}$ and $K_{\sigma}$ are now dependent on this additional parameter $I_{app}$. 
The training set consists of the battery discharge states in time for $I_{app}$ = 10, 12, 14, 15, 16, 18, 20, 22, 24, 25, 27, 28, 29, 30, 32, 33, 34 and 35 A/m$^2$ and
the test set consists of the states for battery discharge at $I_{app}$ = 19 and 31 A/m$^2$. 
Each time instant consists of the 6 battery state variables (see Appendix A, part 5) in 201 nodes (i.e., 1D) and we use the SAK model
to learn the dynamics from one time instant to the next, until the voltage
drops to 3V (also mentioned in Appendix A, part 5). 
We will learn the battery discharge dynamics for discharge current 
$I_{app}$ from the training data and use it to make predictions for the battery discharge for $I_{app}$ values in the test set.
Note that only one model is trained here using all the different $I_{app}$ values in the training set.  
At every iteration during training, a random $I_{app}$ from the training set is considered, from which we sample a random sequence of $n_S$ states of the system.
This random sequence of states (of dimension 201$\times$6) and the $I_{app}$ value are fed as input to the 
conditional SAK model. Since the data is 1D, we use 1D convolutions and deconvolutions. The $I_{app}$ value is
supplied as additional input (i.e., concatenated with the $\mu$ and $\sigma$ of the embedding) to the auxiliary
neural network, which predicts the Koopman matrices $K_{\mu}$ and $K_{\sigma}$
conditioned on $I_{app}$:
\begin{eqnarray}
K_{\mu} (\mu^z_t, \sigma^z_t, I_{app}); \,\,\, K_{\sigma} (\mu^z_t, \sigma^z_t, I_{app}).
\end{eqnarray}
Thus, the Koopman matrices are now generated conditioned on the value of $I_{app}$. For this battery problem, we consider the full $M \times M$ Koopman matrices and 
also use $\lambda_{\mathrm{GAN}}$ = 0.1 (which was obtained after experimentation). 
$n_S$ = 16 at the beginning of the training and is increased by 2.5\% every 20k iteration steps until $n_S$ = 256, after which it is held
fixed. We consider a total of 1 million training iterations, as this problem has more variety in the training data, necessitating longer training. 

\begin{figure}[h]
\centering
(a) \includegraphics[width=6cm]{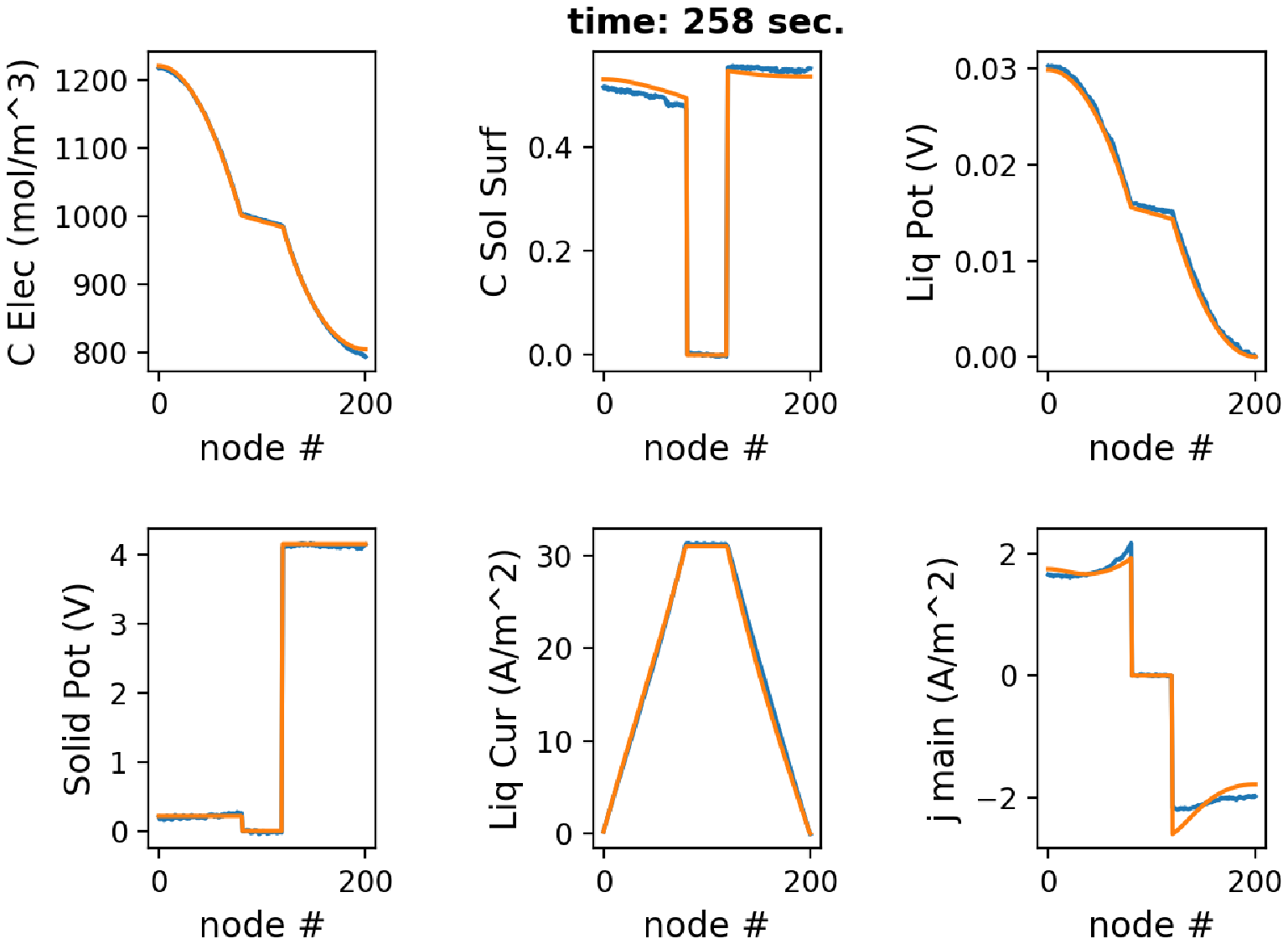} 
(b) \includegraphics[width=6cm]{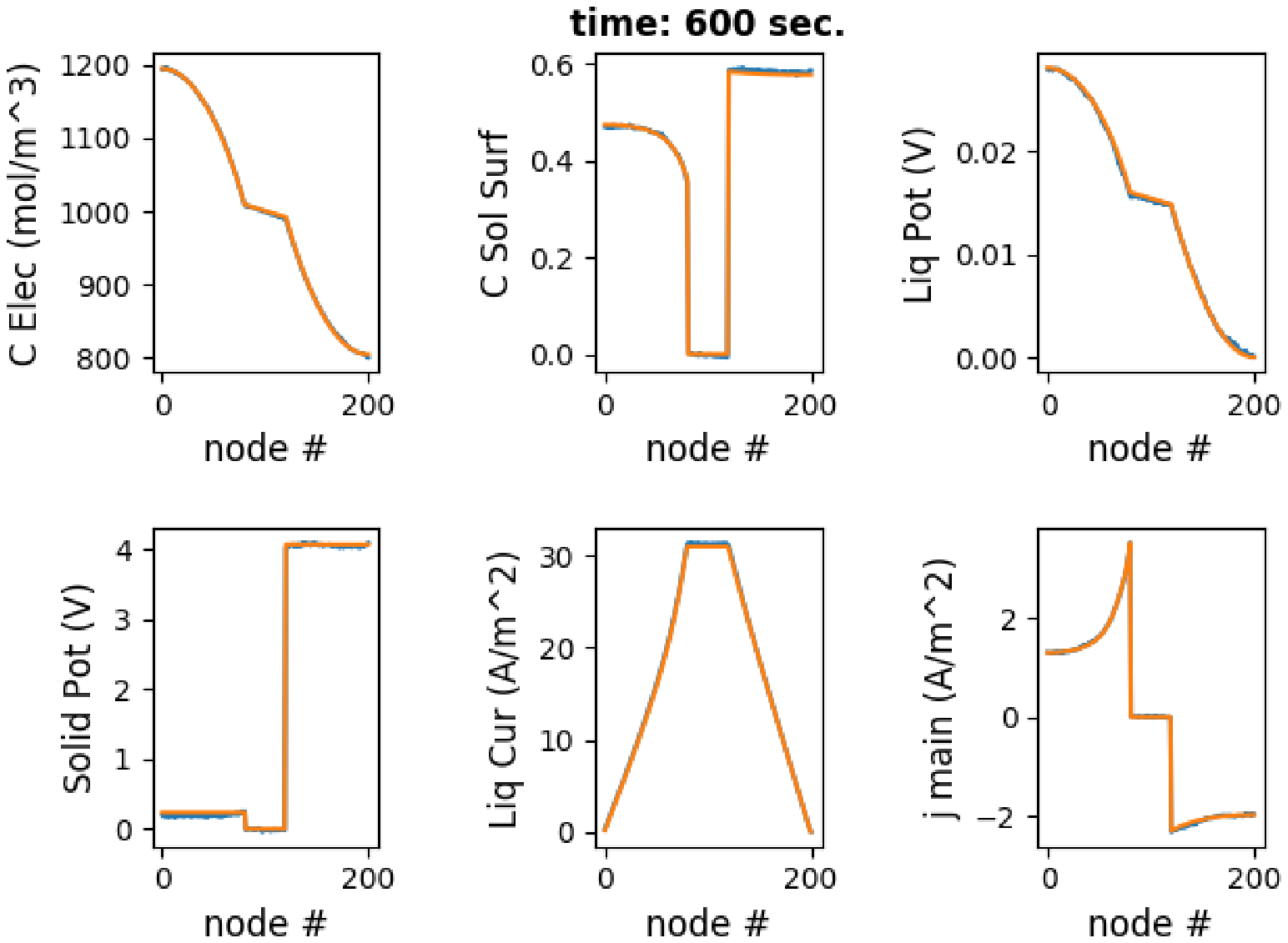} \\
(c) \includegraphics[width=6cm]{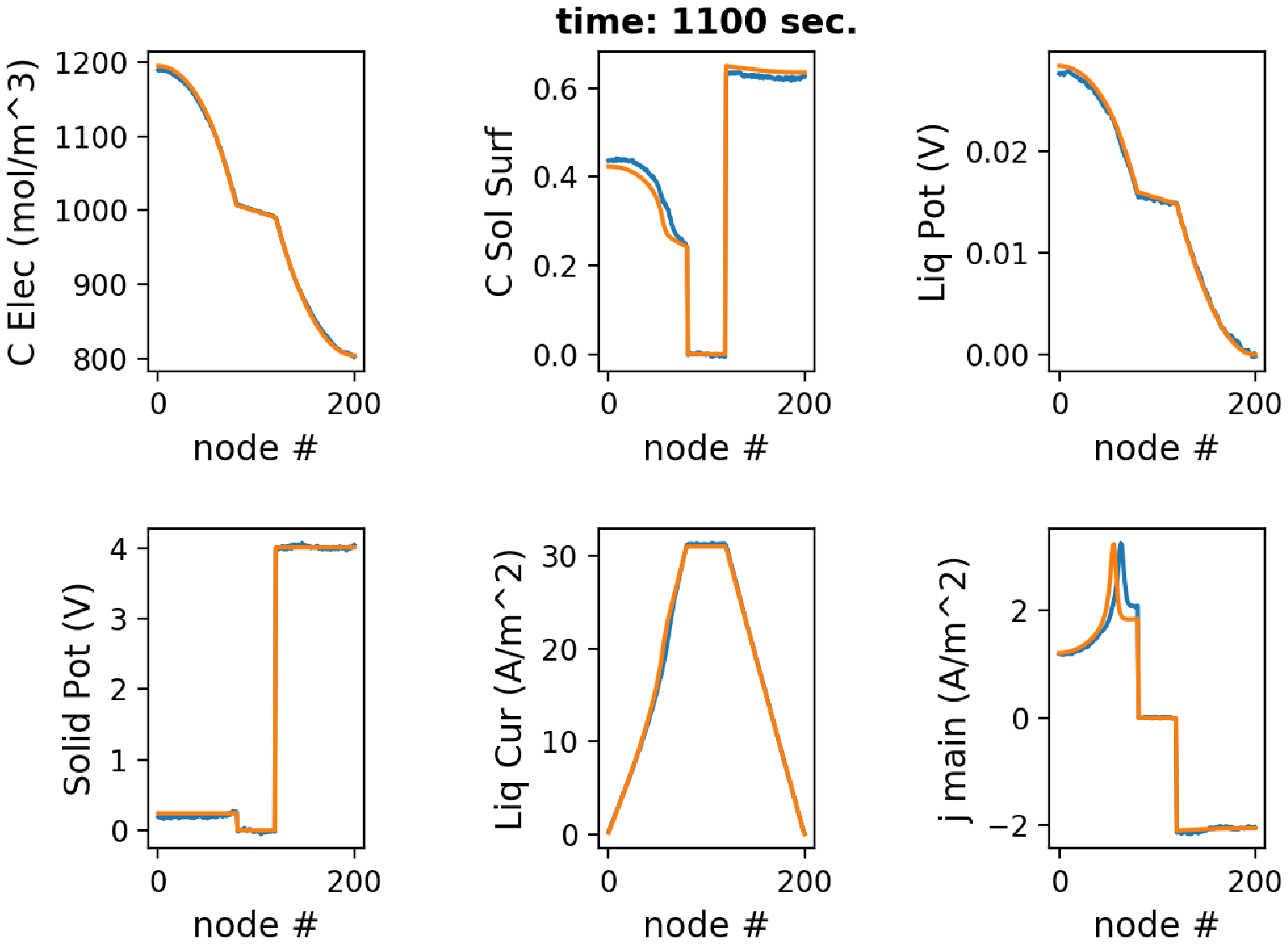} 
(d) \includegraphics[width=6cm]{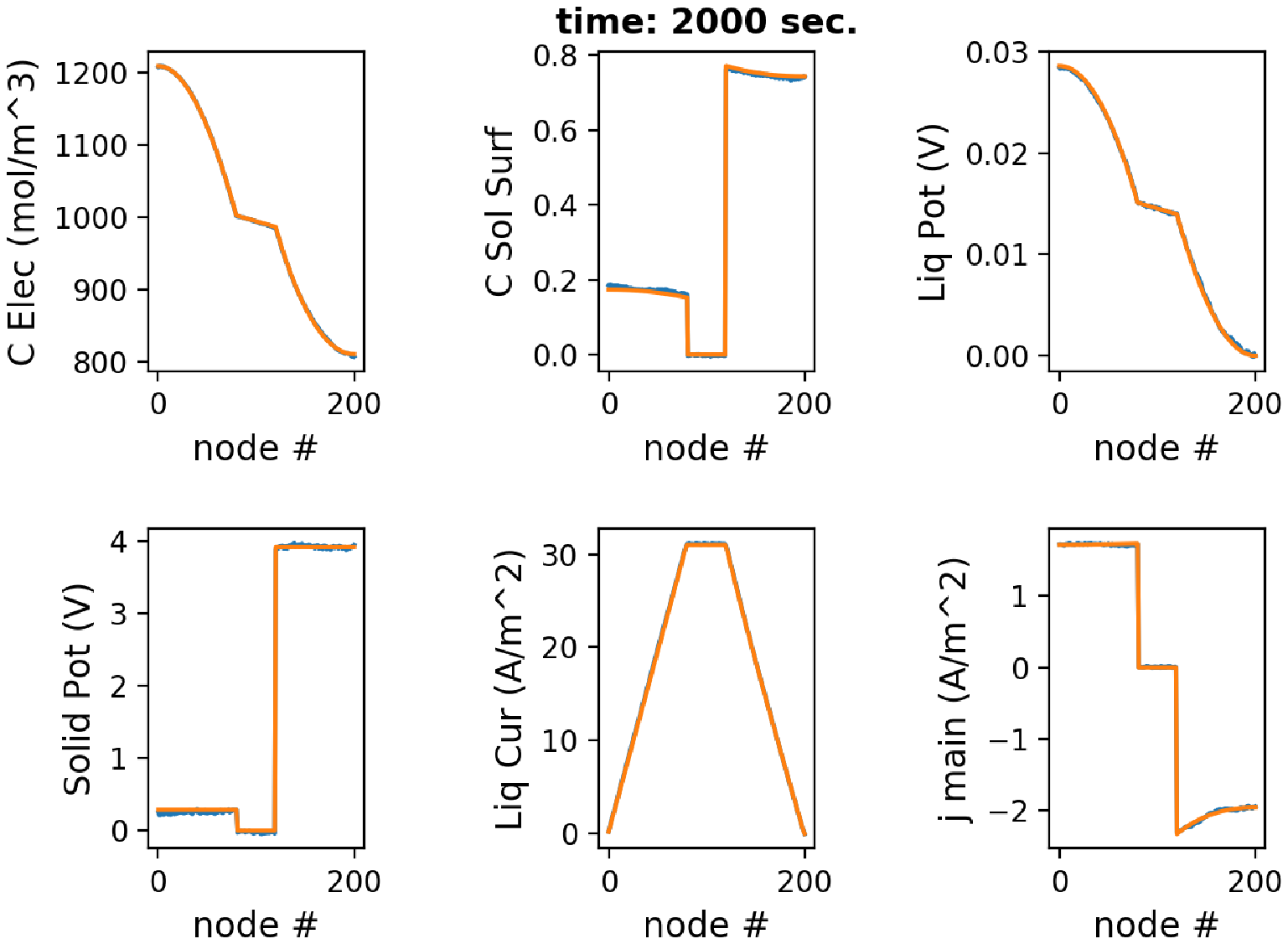} 
\caption{Predictions of the DFN battery model variables for $I_{app}$ = 31 A/m$^2$ with the conditional stochastic adversarial Koopman model at time instants: (a) 258, (b) 600, (c) 1101 and (d) 2000 seconds during discharge. Note that this test case was not part of the training data. The orange curves are the ground truth and the blue curves are the predictions of the conditional SAK model. In some of the figures both the curves overlap and so only one is visible at casual glance. This figure is best viewed by zooming in an electronic version.}
\label{fig:dfn1}
\end{figure}

After the conditional SAK model is trained on the training set, we use it to make predictions for $I_{app}$ values in the test set. 
The model predictions at time instants: 258, 600, 1100 and 2000 seconds during the battery discharge for $I_{app}$ = 31 A/m$^2$ are 
presented in Fig. \ref{fig:dfn1} (in blue), along with the ground truth values (in orange). The model predictions are very accurate, although we do observe
some slight differences for the variables ``C Sol Surf'' and ``j main.'' Thus, the conditional model is able to learn the dynamics of battery discharge for 
certain discharge current density $I_{app}$ values and use it to make predictions for other $I_{app}$ in the test set. 
This conditional flavor is very promising as it has the potential to speed-up the design and development of engineering systems 
which often involve a large number of input parameters. One can generate data for a smaller subspace of design space, use it to train a 
SAK model conditioned on the design parameters, and then make predictions for other values of the design parameters not used in training.

\section{Related work}

Despite the Koopman model \cite{Koopman1931, Koopman1932} being introduced almost a century ago,
in recent years there is a lot of interest to apply the Koopman family
of algorithms to dynamical systems. 
With an increase in the size of data and the availability modern GPU hardware, 
there is a lot of research interest in data-driven modeling of dynamical systems, to which the Koopman models are one prominent family of algorithms \cite{Mezic2013, Takeishi2017, Lusch2018}. 
In particular, the recent interest in Koopman models is to couple 
deep learning methods with classical Koopman models.  
The Dynamic Mode Decomposition (DMD) algorithm \cite{Kutz2016} is one approach to 
represent the Koopman operator. Other deep learning-based approaches such as autoencoders that learn  
to encode from the physical to the Koopman invariant subspace, and vice versa in the decoder, are becoming popular \cite{Takeishi2017, Lusch2018, Morton2018, Balakrishnan20}.
There are different ways of computing the Koopman operator. 
For instance, the Koopman operator was computed by solving a least squares optimization in \cite{Morton2018},
it was evaluated from an auxiliary network and eigenvalue treatment in \cite{Lusch2018} assuming a Jordan block structure for the Koopman matrices.
In \cite{Salova19}, block diagonal structure was used. In \cite{Pan2020}, a tridiagonal Koopman matrix with asymmetric values on either side of the diagonal was used. 
In \cite{Balakrishnan20}, an auxiliary network (similar in spirit to \cite{Lusch2018}) was used to output the full Koopman matrix, including coupling the 
Koopman model with a GAN \cite{GAN}. 
In this study, we extend this approach to a stochastic formulation where the embedding in the Koopman space is modeled as a vector of Gaussian variables and the 
Koopman operator advances this Gaussian variable from one time step to the next.

The deep learning community is constantly developing novel algorithms, some of which can be extended to the study 
of data-driven modeling of dynamical systems. 
In this study, we have demonstrated the efficacy of a Stochastic Adversarial Koopman (SAK) model, including the extension to tridiagonal 
Koopman matrices that operate in the embedding space. 
Furthermore, we have also applied this to a conditional setting where the Koopman operator
is conditioned on additional parameters. 
It is of particular interest to the fluid dynamics community to combine deep learning and Koopman models--as
evidenced by the recent publications \cite{Lusch2018, Morton2018, Pan2020}--as many of these problems are data intensive, which is well suited to train deep learning-based Koopman models. 
Computational finance, epidemology, climate science, etc. are also data intensive and here too Koopman models can play a major role in
furthering our understanding of modeling dynamical systems.


\section{Conclusion}

A novel model based on the Koopman family of algorithms to train dynamical systems is developed in this paper.
The model uses a Gaussian stochastic embedding of an autoencoder as the latent space and advances it in time.
Specifically, an auxiliary network is jointly used to output the Koopman matrices
$K_{\mu}$ and $K_{\sigma}$ that apply on this stochastic embedding to advance it in time. 
Furthermore, the model also couples a GAN discriminator with the stochastic Koopman model and
we term this model as the Stochastic Adversarial Koopman (SAK) model.
The model is robust at learning the dynamics of five different problems that involve partial differential
equations in the governing equations: (1) the Kuramoto-Sivashinsky equation 
for chaos, (2) von Karman vortex shedding behind a cylinder, (3) flame ball-vortex interaction, (4) FitzHugh-Nagumo model,
and (5) the Doyle-fuller-Newman Li-ion battery model.  
The GAN loss is found to robustify the predictions (i.e., lower test errors) in four out of five test problems, provided the right
trade-off parameter $\lambda_{\mathrm{GAN}}$ is used, which requires experimentation and is 
usually problem dependent. A reduced Koopman approach is also considered where the Koopman 
matrices are assumed to be tridiagonal, and this approach produces similar results despite 
having one order of magnitude fewer non-zero entries in the Koopman matrices. 
Finally, we also extend the model to a conditional Koopman model setting where additional input parameters
are supplied to the auxiliary network so that the Koopman matrices are conditioned on these inputs. 
The SAK model predictions are robust and can be used for predicting the dynamics of many real-world systems.

\section*{Acknowledgments}

KB acknowledges the communications with Prof. J. Nathan Kutz of the University of Washington. 

\newpage

\bibliography{StochasticKoopman}

\newpage

\section*{Appendix A: Test Cases}

\subsection*{1. Kuramoto-Sivashinsky equation}

The Kuramoto-Sivashinsky (KS) model is 1D and involves only one variable $u \left(x,t \right)$, whose dynamics is governed by the following fourth-order PDE \cite{Kuramoto1978, Sivashinsky1977}:

\begin{equation}
\frac{\partial u}{\partial t} + u \frac{\partial u}{\partial x} + \frac{\partial^2 u}{\partial x^2} + \frac{\partial^4 u}{\partial x^4} = 0
\end{equation}
in a domain $x \in$ [0, 128] with periodic boundary conditions. The KS model is chaotic and is used to model the diffusive instabilities in a laminar flame front.
The initial condition for KS is given by:

\begin{equation}
u \left(x,0 \right) = \mathrm{cos} \left( x \right) + 0.1 \,\, \mathrm{cos} \left( \frac{x}{16} \right) \,\, \left( 1 + 2\,\, \mathrm{sin} \left( \frac{x}{16} \right) \right).
\end{equation}
The system is solved using Crank-Nicholson/Adams-Bashforth (CNAB2) timestepping \cite{CNAB2} with $\Delta x = \frac{1}{8}$ and $\Delta t = \frac{1}{16}$ for 4800 time steps, with the
solution at every fourth time step saved for the data corpus. Thus, $\Delta t$ = 0.25 for the Koopman analysis and we have a total of 1200 snapshots, with each snapshot
consisting of a vector of 1024 real values (i.e., 128/$\Delta x$).

\subsection*{2. von Karman vortex shedding behind a cylinder}

The classical von Karman vortex shedding behind a cylinder \cite{vonKarman} is widely used for validating many Computational Fluid Dynamics (CFD) codes and is of 
relevance to both the aerospace and automotive industries for aerodynamic drag reduction. The Navier-Stokes equations of fluid dynamics are
solved in a 2D domain for flow past a cylinder from left to right at a Reynolds number $Re$ = 150. The domain size is 22$\times$4 and the velocity of the 
flow at the input is 1 (all units are non-dimensionalized). The domain is discretized using a 660$\times$120 grid and the CFD time step $\Delta t$ = 0.01.   
The open source code: https://github.com/dorchard/navier is used for solving the Navier-Stokes equations and generating the data corpus. 
The flow-field comprises of three variables: $u, v, p$, where $u$ and $v$ are the 
$x$ and $y$ velocity components and $p$ is the pressure.

\begin{figure}[h]
\centering
\includegraphics[width=8cm]{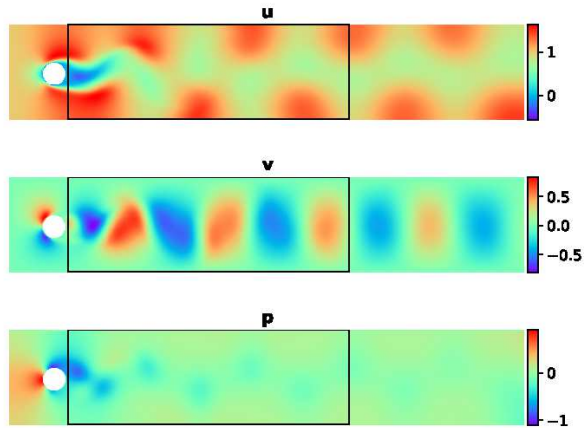} 
\caption{A sample snapshot showing the flow past the cylinder. The cylinder is the white circle and the flow is from left to right. The three sub-figures show the velocity components $u$ and $v$ in the $x$- and $y$-directions, respectively, and the pressure $p$. The black rectangle shows the region from the Navier-Stokes solver that is used for training the Deep Adversarial Koopman models.}
\label{fig:karmanFull}
\end{figure}

For training the SAK model, only the region immediately behind the cylinder where the vortex shedding is preponderant, is considered; specifically, we consider
360 cells in the horizontal immediately behind the cylinder and the entire 120 cells in the vertical directions.
Every 4-th CFD solution snapshot is saved, and thus $\Delta t$ = 0.04 for the Koopman analysis.
We consider 400 time snapshots for training the SAK model, which corresponds to approximately 3.5 cycles of vortex shedding. 
A sample snapshot is presented in Fig. \ref{fig:karmanFull} for illustration.

\subsection*{3. Flame ball-vortex interaction}

Flame-vortex interaction is a classical engineering problem in combustion engines, where vortices can distort a flame, thereby enhancing the mixing of fuel with the oxidizer. 
We consider a simple 2D model \cite{Roussel05} for dimensionless temperature $T$ and mass fraction/concentration of premixed gas $Y$
and this consists of a system of 2 PDEs. For brevity, these equations are not presented and can be found in \cite{Roussel05}. 
Outflow boundary conditions are used at all four boundaries for the CFD analysis. 
The initial conditions are are not presented here and can be found in \cite{Roussel05}. 
We use a 2$^{nd}$ order central scheme in space and a 3$^{rd}$ order Runge-Kutta scheme in time to solve the system of equations. 
For the CFD analysis, the domain size is 20$\times$20 and is discretized using a 512$\times$512 mesh for the finite difference method, with a time step $\Delta t$ = 10$^{-5}$ for 140k time steps. 
The solution at every other node is saved once every 1000 time steps, and so the data corpus for training the SAK model is
of size 256$\times$256$\times$2 (the 2 is for two variables: $T,Y$) for 141 snapshots.

\subsection*{4. FitzHugh-Nagumo model}

The next problem we consider is the FitzHugh-Nagumo model \cite{FitzHugh, Carletti20}, which 
is a classical problem used for modeling nerve impulses as well as Turing patterns \cite{Turing1952}.
The model consists of two variables $u$ and $v$ which are solved as:
\begin{eqnarray}
\frac{\partial u}{\partial t} = a \left( \frac{\partial^2 u}{\partial x^2} + \frac{\partial^2 u}{\partial y^2} \right) + u - u^3 - v + k, \nonumber \\ 
\frac{\partial v}{\partial t} = \frac{1}{\tau} \left[ b \left( \frac{\partial^2 v}{\partial x^2} + \frac{\partial^2 v}{\partial y^2} \right) + u - v \right]. \nonumber \\ 
\end{eqnarray}
The values of the constants are $a$ = 2.8$\times$10$^{-4}$, $b$ = 5$\times$10$^{-3}$, $\tau$ = 0.1 and k = -5$\times$10$^{-3}$. A finite difference method is
used to first generate the data corpus. For this a 64$\times$64 mesh is used and the 2D domain size is $0 \le (x,y) \le 1$. 
Both $u$ and $v$ are randomly initialized and Neumann boundary condition are applied at all boundaries (i.e., first derivatives of $u$ and $v$ set to zero). 
The finite difference time step $\Delta t$ = 0.001 and the model is run till time instant of 10 (i.e., 10k time steps).
The solution snapshot is saved once every 50 time steps, and so the effective time step between each 
snapshot in the data corpus is 0.05. Note that we start with a random initialization for this problen, and since a neural network cannot be trained 
predict random values, we use the data corpus for time instants $>$ 1.0 only, so that a few patterns have already started to emerge in the data.   
We will use this data to train the SAK model, and since we train from time instants 1.0--10.0 in increments of 0.05, we have a
total of 180 snapshots in the data corpus.

\subsection*{5. Doyle-Fuller-Newman Li-ion battery model}
\label{sec:dfn}

The Doyle-Fuller-Newman (DFN) model consists of a system of PDEs in 1D to solve for the Li-ion concentrations and potential in the electrode particles and the electrolyte of a Li-ion
battery. The equations are not presented here for brevity and the interested reader is referred to \cite{Dualfoil1, Dualfoil2}. We use the  
Dualfoil code v5.1 open sourced from the DFN authors \cite{Dualfoil3} for generating the data corpus. Specifically, we consider a fully charged Li-ion battery and 
vary the applied current density, $I_{app}$, in the range 10-35 A/m$^2$, and make predictions of the battery potentials in the electrode and electrolyte, the 
Li-ion concentrations, and the current as the batetry discharges in time. The cross-section of the battery is discretized using 201 nodes and the Dualfoil code 
outputs 6 variables: concentration of the Li-ions in the electrolyte (in mol/m$^3$), concentration of Li-ions at the surface of the solid particles (non-dimensionalized),
potentials in the liquid and solid phases (both in Volts), liquid phase current density and current density ``j main" (both in A/m$^2$). 
These variables are saved every 1 second time interval during the discharge of the battery until the voltage drops to 3 V, and the total number of time snapshots
varies depending on $I_{app}$; for instance, the total number of snapshots is 8454 for $I_{app}$ = 10 A/m$^2$ and is 2197 for $I_{app}$ = 35 A/m$^2$, with the other cases in between.        
More details on the battery modeling system and solution procedures can be found in the Dualfoil manual \cite{Dualfoil3}.

We train the SAK model with the Koopman matrix conditioned on $I_{app}$ by supplying this as an input to the auxiliary network. 
The SAK model is trained on a training set comprising of 18 different $I_{app}$ values in the range 10 $\le I_{app} \le$ 35 A/m$^2$ and used to make predictions
of the state of the battery at different time instants for a value of $I_{app}$ not used
in the training. 


\section*{Appendix B: Neural Network Architectures}

The neural network architecture used in the analysis is summarized here. 
Note that we have a total of 4 neural networks: \textit{Encoder}, \textit{Decoder}, auxiliary network \textit{AUX} (to obtain $K$ matrices) and the GAN discriminator \textit{DISC}.  
We will use several different deep learning building blocks: batch normalization (\textit{BN}) \cite{BN}, Dropout (\textit{Dropout}) \cite{dropout},
convolutional (\textit{conv}) and deconvolutional (\textit{dconv}) \cite{dconv} operators, and the Relu (\textit{Relu}) activation function.
The notation \textit{conv}(k,f,S,s) is used for a convolutional layer with kernel size $k$, $f$ filters, same padding (identified by $S$) and a stride of $s$. 
The notation \textit{Dense}($n$) is used to refer to a fully connected dense layer with $n$ neurons. 
We first define a bottleneck layer for \textit{Encoder} with $N_f$ filters as input, \textit{BottleNeck}$^{e}$($N_f$), comprising of the following in the same order:    
\textit{BottleNeck}$^{e}$($N_f$) = \textit{BN} $\rightarrow$ \textit{Relu} $\rightarrow$  \textit{conv}(1,$N_f$/2,S,1) $\rightarrow$  \textit{BN} $\rightarrow$  \textit{Relu} $\rightarrow$  \textit{conv}(3,$N_f$/2,S,1) $\rightarrow$  \textit{BN} $\rightarrow$  \textit{Relu} $\rightarrow$  \textit{conv}(1,$N_f$,S,1).
\textit{Encoder} consists of 5 layers of convolutional operations supplemented with bottleneck layers added residually, similar to Resnet \cite{Resnet}. 
For ease of notation, we will refer to \textit{Encoder}'s residual block as \textit{RES}$^{e}$($N_f$) = \textit{conv}(3,$N_f$,S,2) + \textit{BottleNeck}$^{e}$($N_f$).
Specifically, \textit{Encoder} consists of 5 residual layers in succession: \textit{RES}$^{e}$(64) $\rightarrow$ \textit{RES}$^{e}$(128) $\rightarrow$ \textit{RES}$^{e}$(256) $\rightarrow$
\textit{RES}$^{e}$(512) $\rightarrow$ \textit{RES}$^{e}$(512), followed by a \textit{Relu} and a flattening operation. This is then fed into two \textit{Dense}($M$) layers to
output the $\mu$ and $\sigma$ of the embedding vector, where the dimension $M$ = 64 is used throughout this study. 

For the \textit{Decoder} we define a similar bottleneck layer with $N_f$ filters, albeit this time using deconvolutional operations:  
\textit{BottleNeck}$^{d}$($N_f$) = \textit{BN} $\rightarrow$ \textit{Relu} $\rightarrow$ \textit{dconv}(1,$N_f$/2,S,1) $\rightarrow$ \textit{BN} $\rightarrow$ \textit{Relu} $\rightarrow$ \textit{dconv}(3,$N_f$/2,S,1) $\rightarrow$ \textit{BN} $\rightarrow$ \textit{Relu} $\rightarrow$ \textit{dconv}(1,$N_f$,S,1). 
For \textit{Decoder}, we first add the input to the bottleneck layer akin to
Resnet \cite{Resnet}, which is then passed though a deconvolutional layer, like so: \textit{RES}$^{d}$($N_f$) = input + \textit{BottleNeck}$^{d}$($N_f$) $\rightarrow$ \textit{dconv}(3,$N_f$,S,2).
\textit{Decoder} starts with \textit{Dense}($\cdot$) with the number of neurons used being the same as the dimension of the \textit{Encoder}'s flattened output. 
This is reshaped as appropriate and is followed by 5 layers of the \textit{Decoder}'s residual blocks in succession: \textit{RES}$^{d}$(512) $\rightarrow$ \textit{RES}$^{d}$(256) $\rightarrow$ \textit{RES}$^{d}$(128) $\rightarrow$ \textit{RES}$^{d}$(64) $\rightarrow$ \textit{RES}$^{d}$($n_{out}$). Here, $n_{out}$ is the number of output channels in the data. 
Note that some problems are 1D whereas others are 2D, and so the appropriate \textit{conv} and \textit{dconv} API calls are used. 

For the \textit{AUX} network, we define a fully connected layer with $N$ neurons as \textit{FC}($N$) = \textit{Dense}($N$) $\rightarrow$ \textit{Relu} $\rightarrow$ \textit{Dropout}. 
For $\textit{Dropout}$, we set the probability of keeping the activations to 0.8 at training, and 1.0 at testing.   
\textit{AUX} network consists of 4 fully connected layers: \textit{FC}(128) $\rightarrow$ \textit{FC}(256) $\rightarrow$ \textit{FC}(512) $\rightarrow$ \textit{Dense}(2$n_{dec}$) 
(the 2 factor arises since we consider two Koopman matrices $K_{\mu}$ and $K_{\sigma}$).
For the full Koopman matrices, $n_{dec}$ = $M^2$, whereas for the tridiagonal Koopman matrices, $n_{dec}$ = 3$M$-2.   

For \textit{DISC}, we will use the Leaky Relu activation function, denoted as \textit{LRelu}, with a slope of 0.2 in the negative side.
We define a block \textit{B}$^\mathrm{DISC}$($N_f$) as \textit{conv}(5,$N_f$,S,2) $\rightarrow$ \textit{BN} $\rightarrow$ \textit{LRelu}.
\textit{DISC} is then constructed as: \textit{conv}(5,64,S,2) $\rightarrow$ \textit{LRelu} $\rightarrow$ \textit{B}$^\mathrm{DISC}$(128) $\rightarrow$ \textit{B}$^\mathrm{DISC}$(256) $\rightarrow$ \textit{B}$^\mathrm{DISC}$(512). The output is then reshaped and passed to a \textit{Dense}(1) without any activation function to represent the Wasserstein distance. 
 
Adam \cite{Adam} optimizer is used to train the neural networks with a learning rate of 1$\times$10$^{-5}$.
The total number of iterations used for training varies for the different problems, but is usually in the order of 100k-500k, where at each iteration step one sequence
of $n_{S}$ contiguous snapshots are randomly sampled from the data corpus and used to train the networks. 


\end{document}